\documentclass{article} 
\usepackage[table]{xcolor}
\usepackage[preprint]{colm2025_conference}

\usepackage{microtype}
\usepackage{hyperref}
\usepackage{url}
\usepackage{booktabs}

\usepackage{lineno}
\usepackage{graphicx}
\usepackage{amsmath,lipsum}

\usepackage{CJKutf8}

\usepackage{booktabs}
\usepackage{multirow}
\usepackage{adjustbox}
\usepackage{wrapfig}
\usepackage{pifont}
\usepackage{subcaption}
\usepackage{tcolorbox}

\definecolor{darkblue}{rgb}{0, 0, 0.5}
\hypersetup{colorlinks=true, citecolor=darkblue, linkcolor=darkblue, urlcolor=darkblue}

\title{Evaluating Large Language Model with Knowledge Oriented Language Specific Simple Question Answering}

\author{
\textbf{Bowen Jiang\textsuperscript{2}\thanks{Equal contribution.}}~,
\textbf{Runchuan Zhu\textsuperscript{2}\footnotemark[1]}~,
\textbf{Jiang Wu\textsuperscript{1}\footnotemark[1]~~\thanks{Project lead.}}~,
\textbf{Zinco Jiang\textsuperscript{2}},
\textbf{Yifan He\textsuperscript{2}},\\
\textbf{Junyuan Gao\textsuperscript{3}}, 
\textbf{Jia Yu\textsuperscript{1}},
\textbf{Rui Min\textsuperscript{1}},
\textbf{Yinfan Wang\textsuperscript{4}},
\textbf{Haote Yang\textsuperscript{1}}, \\
\textbf{Songyang Zhang\textsuperscript{1}},
\textbf{Dahua Lin\textsuperscript{1}},
\textbf{Lijun Wu\textsuperscript{1}},
\textbf{Conghui He\textsuperscript{1}\thanks{Corresponding author.}} \\
\textsuperscript{1}Shanghai Artificial Intelligence Laboratory \\
\textsuperscript{2}Peking University \\
\textsuperscript{3}University of Chinese Academy of Sciences \\
\textsuperscript{4}Shanghai University \\
\texttt{heconghui@pjlab.org.cn}
}

\begin{document}
\begin{CJK}{UTF8}{gbsn}

\ifcolmsubmission
\linenumbers
\fi

\maketitle

\begin{abstract}
We introduce KoLasSimpleQA, the first benchmark evaluating the multilingual factual ability of Large Language Models (LLMs). Inspired by existing research, we created the question set with features such as single knowledge point coverage, absolute objectivity, unique answers, and temporal stability. These questions enable efficient evaluation using the LLM-as-judge paradigm, testing both the LLMs'  factual memory and self-awareness  (``know what they don't know'').
KoLasSimpleQA expands existing research in two key dimensions:
(1) \textbf{Breadth (Multilingual Coverage)}: It includes 9 languages, supporting global applicability evaluation.
(2) \textbf{Depth (Dual Domain Design)}: It covers both the general domain (global facts) and the language-specific domain (such as history, culture, and regional traditions) for a comprehensive assessment of multilingual capabilities.
We evaluated mainstream LLMs, including traditional LLM and emerging Large Reasoning Models. Results show significant performance differences between the two domains, particularly in performance metrics, ranking, calibration, and robustness. This highlights the need for targeted evaluation and optimization in multilingual contexts.
We hope KoLasSimpleQA will help the research community better identify LLM capability boundaries in multilingual contexts and provide guidance for model optimization.
We will release KoLasSimpleQA at \url{https://github.com/opendatalab/KoLasSimpleQA} .
\end{abstract}

\section{Introduction}
\label{sec:Introduction}

Large Language Models (\cite{grattafiori2024llama, yang2024qwen2, guo2025deepseek}) have advanced significantly, yet hallucination—where models produce unverified or misleading information—remains a major challenge, affecting their reliability and broader use. To tackle this, the SimpleQA (\cite{wei2024measuring}) and ChineseSimpleQA (\cite{he2024chinese}) benchmarks were introduced to evaluate LLMs' factual ability using short, fact-based questions. These questions focus on a single knowledge point, with answers that are objective, stable, and not open to interpretation.

Recent studies (\cite{zhang2023don, shi2022language, huang2023not}) indicate that LLM performance varies across languages, particularly in factual ability, with models typically performing better in English. However, the SimpleQA and ChineseSimpleQA benchmarks are limited to English and Chinese. Additionally, most evaluations of LLMs in non-English contexts focus on general knowledge rather than language-specific content like history, culture, and local traditions. While LLMs excel in general knowledge, they often struggle with language-specific facts.

\begin{table}[htbp]
    \centering
     \begin{tabular}{c|c|ccc}
\hline
Benchmark       & Lang.                    & Lang. specific  & Easy to evaluate \\ \hline
XTREME (\cite{hu2020xtreme})          & 40 lang. & \ding{55}                    & \ding{55}        \\
Okapi (\cite{lai2023okapi})           & en        & \ding{55}            & \ding{55}        \\
SimpleQA (\cite{wei2024measuring})        & en                       & \ding{55}                         & \ding{51}        \\
ChineseSimpleQA (\cite{he2024chinese}) & zh                       & \ding{55}                          & \ding{51}        \\
MINTQA (\cite{he2024mintqa})          & en                       & \ding{55}                         & \ding{55}        \\
BenchMAX (\cite{huang2025benchmax}        & 17 lang.     & \ding{55}                       & \ding{55}        \\
MMLU-ProX (\cite{xuan2025mmlu})       & 42 lang.     & \ding{55}                       & \ding{51}        \\ \hline
\textbf{KoLasSimpleQA(Ours)}   & 9 lang.      & \ding{51}                  & \ding{51}        \\ \hline
\end{tabular}
    \caption{Comparison between KoLasSimpleQA and other benchmarks. The comparison is conducted along three dimensions: the range of supported languages, whether the benchmark includes language-specific knowledge, and whether it is easy to evaluate.}
  \label{tab:dataset}%
\end{table}%

To tackle this problem, we introduce \textbf{K}nowledge-\textbf{O}riented \textbf{La}nguage-\textbf{S}pecific \textbf{Simple} \textbf{Q}uestion \textbf{A}nswering (KoLasSimpleQA), a benchmark comprising simple fact-based QA samples grounded in genuinely language-specific knowledge across nine languages.
KoLasSimpleQA has three main features:
\textbf{(1) Foundation:} Inspired by~\cite{wei2024measuring,he2024chinese}, we crafted a question set with attributes such as \textit{single knowledge point coverage}, \textit{absolute objectivity}, \textit{unique answers}, and \textit{temporal stability}. These questions enable efficient evaluation using the LLM-as-judge paradigm, assessing both the factual memory and self-awareness of LLMs (i.e., their ability to "know what they don't know").
\textbf{(2) Breadth Expansion:} Unlike existing work~\cite{wei2024measuring,he2024chinese} that is limited to a few languages, KoLasSimpleQA includes \textit{9 languages}, allowing for performance evaluation of LLMs in multilingual contexts and supporting assessments of global applicability.
\textbf{(3) Depth Exploration:} It encompasses both the \textit{general domain} (global facts) and the \textit{language-specific domain} (such as history, culture, and regional traditions). We collected data from Wikipedia, categorizing entries into general and language-specific domains based on the number of inter-language links each article has (as shown in Figure~\ref{fig:ill}; see details in \S\ref{sec:collection_of_specific_and_general_entry}). Using these classifications and filtered entries, we constructed questions for both domains, enabling a comprehensive and in-depth evaluation of LLMs in multilingual settings.

We evaluated mainstream LLMs on KoLasSimpleQA, covering both traditional LLMs and the latest Large Reasoning Models (LRMs). Key insights from our study include:

\textbf{(1) Performance Disparity Across Domains:} LLMs perform much worse in the language-specific domain than in the general domain.

\textbf{(2) Translating Non-English Queries:} Translating queries into English is a common strategy to enhance multilingual performance. While effective in the general domain, it is less so in the language-specific domain.

\textbf{(3) Calibration Performance:} LLMs show significantly poorer calibration in the language-specific domain compared to the general domain.

\textbf{(4) Knowledge Memorization Robustness:} LLMs are notably less robust in the language-specific domain than in the general domain.
\section{KoLasSimpleQA}
\label{sec:KoLasSimpleQA}

\subsection{Overview} 

\begin{figure}[t]
  \centering
  \includegraphics[width=\linewidth]{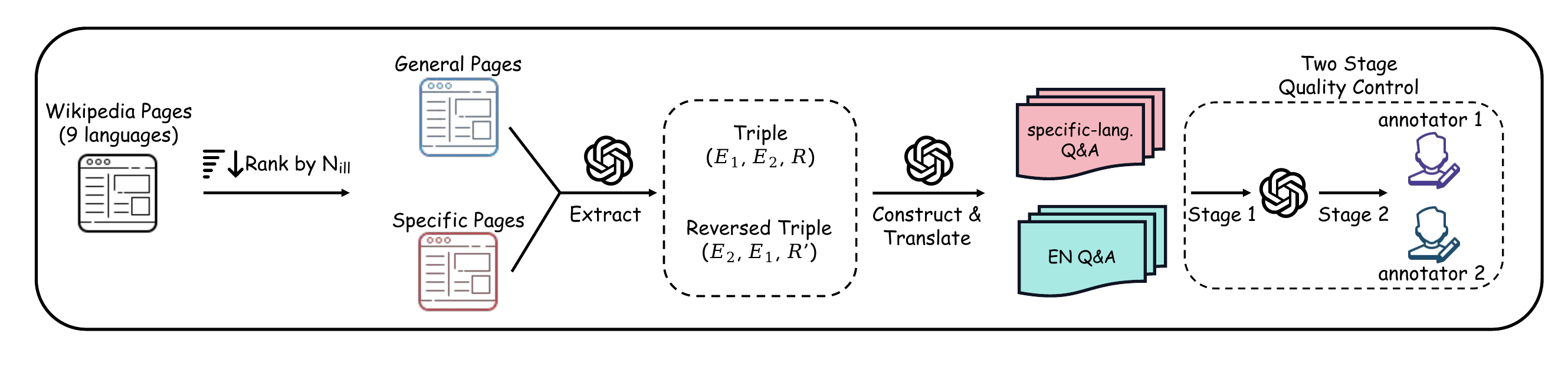}
  \caption{Construction pipeline of KoLasSimpleQA. The process includes Wikipedia entry selection based on inter-language links, triple and QA pair generation using GPT-4o, and a two-stage quality control to ensure question quality and diversity.}
  \label{fig:construction}
\end{figure}

We created a multilingual QA benchmark including both the general and language-specific domains. The general domain covers global knowledge shared across languages, while the language-specific subset targets knowledge unique to individual linguistic and cultural contexts. Our benchmark includes 9 languages: Hungarian (hu), Czech (cs), Serbian (sr), Russian (ru), Chinese (zh), Korean (ko), Thai (th), Arabic (ar), and Vietnamese (vi), along with their English (en) translations. The data construction process is shown in Figure \ref{fig:construction}.

\subsection{Benchmark Construction}
\subsubsection{Collection of Specific \& General Wikipedia Entries}
\label{sec:collection_of_specific_and_general_entry}

\begin{figure}[t]
  \centering
  \includegraphics[width=1.0\linewidth]{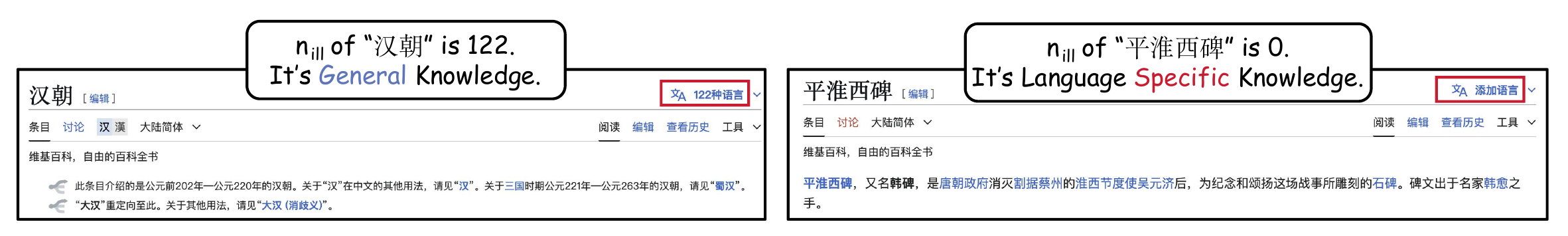}
  \caption{Illustration of inter-language links on a Wikipedia page. The number of such links (\( n_{\text{ill}} \)) is used to distinguish between language-specific and language-general knowledge.}
  \label{fig:ill}
\end{figure}

We crawled all Wikipedia pages in nine languages and extracted their contents. Unlike conventional methods focusing only on entry content, we specifically extracted inter-language link information for each entry and counted the number of these links, denoted as \( n_{\text{ill}} \).
An inter-language link is a hyperlink that connects a Wikipedia page in one language to the related page in another language, aiding navigation across different language versions. These links are usually located in the sidebar at the top right of the webpage, as shown in Figure~\ref{fig:ill}.
It's important to note that pages in different language versions of Wikipedia are not machine-translated; they are created by users who are typically native speakers of the respective language. If a page exists in a particular language, the topic or knowledge it represents is likely of interest to native speakers of that language.
Thus, if an entity lacks versions in other languages (\( n_{\text{ill}} = 0 \)), it is classified as \textbf{language-specific}, indicating the knowledge is unique to that language. Conversely, if an entity appears in many language versions (\( n_{\text{ill}} \) is large), it likely represents \textbf{global knowledge}, making it \textbf{general}.

For each of the nine languages, we sample a total of \( N_{\text{specific}} + N_{\text{general}} \) entries. Specifically, we randomly select \( N_{\text{specific}} \) entries from those where \( n_{\text{ill}} = 0 \), ensuring that they represent language-specific knowledge. Formally, this subset is defined as \( \mathbf{E_{\text{specific}}^{\text{selected}}} = \{ x \in \mathbf{D} \mid n_{\text{ill}}(x) = 0, |\mathbf{E_{\text{specific}}^{\text{selected}}}| = N_{\text{specific}} \} \). Additionally, we identify the top \( N_{\text{general}} \) entries with the highest \( n_{\text{ill}} \) values, as these are more likely to contain language-general knowledge, defined as \( \mathbf{E_{\text{general}}^{\text{selected}}} = \{ x \in \mathbf{D} \mid \texttt{Rank}(n_{\text{ill}}(x)) \leq N_{\text{general}} \} \).
Here, \( \mathbf{D} \) represents the set of all Wikipedia entries, \( \mathbf{E_{\text{specific}}^{\text{selected}}} \) and \( \mathbf{E_{\text{general}}^{\text{selected}}} \) denote the selected sets of language-specific and language-general entries, respectively. The function \( n_{\text{ill}}(x) \) returns the number of inter-language links associated with entity \( x \), and \( \texttt{Rank}(n_{\text{ill}}(x)) \) represents the ranking of entity \( x \) based on its \( n_{\text{ill}} \) value. Finally, \( N_{\text{specific}} \) and \( N_{\text{general}} \) define the number of sampled language-specific and language-general entries, respectively. We set \( N_{\text{specific}} = 400 \) and \( N_{\text{general}} = 400 \).

\subsubsection{Construction of Question-Answer}

We employ GPT-4o to extract triples (E1, R, E2) that satisfy the requirements from the collected general and specific Wikipedia entries (See prompt in Table~\ref{tab:prompt_extract_triple}). Subsequently, we leverage GPT-4o to generate Question-Answer (QA) pairs from these triples and translate them into their corresponding English parallel QA pairs (see prompt in Table~\ref{tab:prompt_construct_qa_from_triple} and~\ref{tab:prompt_translate_non_en_to_en}. For example, if the title of a Wikipedia entry is ``Lombardy,'' and one segment of the content is ``Lombardy is located between the Alps mountain range and tributaries of the river Po, and includes Milan, its capital, the largest metropolitan area in the country, and among the largest in the EU,'' the extracted triple would be (Lombardy, capital is, Milan), and the corresponding QA pair would be ``What is the capital city of the Lombardy region?''

Additionally, we extract the reversed triples (E2, R', E1) and construct QA pairs that explicitly capture inverse relations. This approach allows us to assess whether the model has genuinely internalized the underlying knowledge, as discussed in this study (\cite{berglund2023reversal}). The reversed triple corresponding to the above example is (Milan, is capital of, Lombardy), and the corresponding QA pair would be ``What Italian region is Milan the capital of?''

\begin{table}[t]
\begin{center}
\begin{tabular}{c|ccccccccc|c}
\hline
 & ar & cs & hu & ko & ru & sr & th & vi & zh & total \\
\hline
\textbf{General Domain} & 162 & 142 & 169 & 151 & 92 & 133 & 209 & 121 & 127 & 1306 \\
\textbf{Language Specific Domain} & 117 & 90 & 127 & 84 & 57 & 79 & 99 & 92 & 96 & 841 \\
\hline
\end{tabular}
\end{center}
\caption{Distribution of question counts across languages in KoLasSimpleQA.}
\label{sample-table}
\end{table}

\begin{figure}[t]
  \centering
  \includegraphics[width=1.0\linewidth]{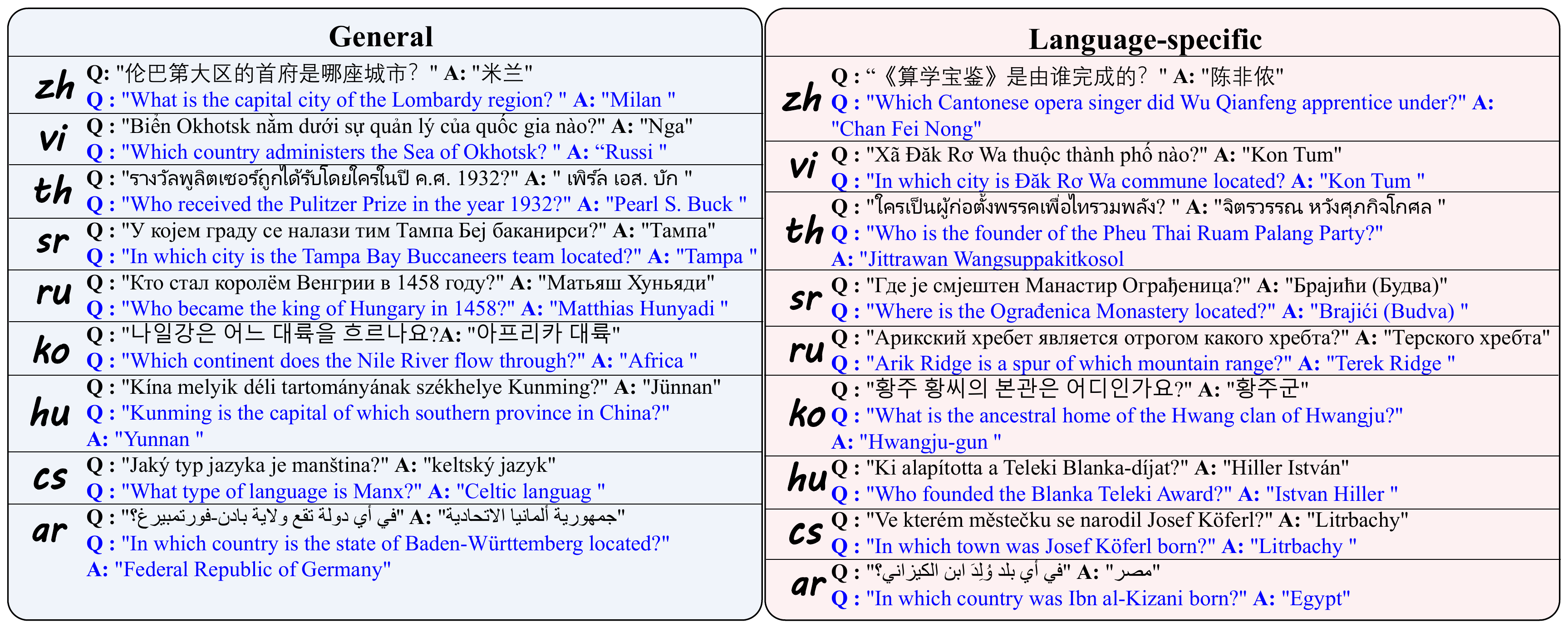}
  \caption{Example QA pairs in KoLasSimpleQA.}
  \label{fig:Kolas examples}
\end{figure}

\begin{figure}[!t]
  \centering
  \includegraphics[width=1.0\linewidth]{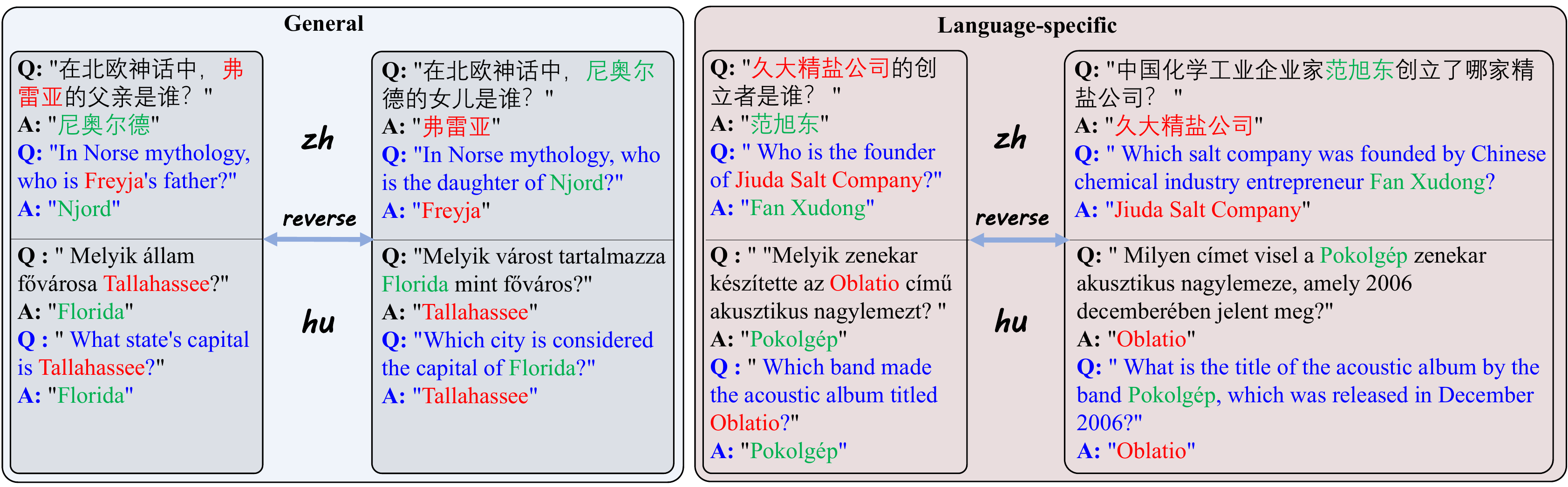}
  \caption{Example QA pairs in the reverse relationship in KoLasSimpleQA.}
  \label{fig:Kolas examples reverse}
\end{figure}

\subsubsection{Two-stage Quality Control}

For each constructed QA pair, we expect them to meet the following criteria:
\textbf{(1) Triple Consistency (TC)}: The question must be generated from the head entity and relation of the triple, while the answer must correctly correspond to the tail entity.
\textbf{(2) Self-Sufficiency (SS)}: The question should contain all necessary information to be answerable on its own, without requiring external context.
\textbf{(3) Non-Triviality (NT)}: The question should not be overly simplistic, and the answer should not be directly inferable from the question itself.
\textbf{(4) Objectivity (OBJ)}: The question must be based on verifiable, factual knowledge, avoiding opinion-based, subjective, or interpretive content.
\textbf{(5) Temporal Stability (TS)}: The answer must remain valid over time and should not be subject to change.
\textbf{(6) Answer Uniqueness (AU)}: The question must be specific enough to elicit a single, unambiguous answer. Avoid vague formulations like ``when'' or ``where'' that may lead to multiple valid answers; instead, use precise expressions such as ``which year'' or ``which city''.

To ensure questions meet standards, we adopted a two-stage Quality Control strategy.
In Stage 1, GPT-4o filters the questions based on the predefined criteria (see prompt in Table~\ref{tab:prompt_quality_control_stage1}).
In Stage 2, native-speaking human annotators review the questions without reference answers, checking if they meet the criteria. For qualified questions, annotators provide answers, using search engines if needed, and cite sources (e.g., URLs). The LLM-as-judge method then compares the annotators' answers with the reference answer (see Table \ref{tab:llm_as_judge_3}). If both annotators agree on the correctness of the reference answer, the question is deemed qualified.
Finally, to ensure the diversity of the questions, we further filtered the questions from the same entry, ensuring that each entry corresponds to only one QA pair (QA pairs in a reverse relationship are excluded).

\subsection{Benchmark Summary}

Throughout the construction and quality control of KoLasSimpleQA, a large number of low-quality question–answer pairs were filtered out. Initially, 400 Wikipedia entries were selected for each language. After a rigorous two-stage quality control process, a total of 2,147 QA pairs were retained across all languages. Among these, there are 452 reverse relationship QA pairs in the general domain and 190 in the language-specific domain.
The distribution of QA pairs per language is shown in Table~\ref{sample-table}. Representative examples of QA pairs in KoLasSimpleQA are presented in Figure~\ref{fig:Kolas examples}, while examples of QA pairs in a reverse relationship are illustrated in Figure~\ref{fig:Kolas examples reverse}.

\section{Experiment}
\label{sec:Experiment}

\subsection{Experimental setup}
We evaluated a selection of state-of-the-art large language models (LLMs), including GPT-4o~\cite{hurst2024gpt} and GPT-4o-mini\footnote{We used \texttt{gpt-4o-2024-11-20} for GPT-4o and \texttt{gpt-4o-mini-2024-07-18} for GPT-4o-mini.}, as well as Deepseek-V3~\cite{liu2024deepseek}, Qwen2.5-Instruct~\cite{yang2024qwen2}, and Llama-3.1-Instruct~\cite{wendler2024llamas}. Additionally, we assessed the latest Large Reasoning Models (LRMs), including OpenAI o1-mini, QwQ-32B-Preview, QwQ-32B (abbreviated as QwQ-preview and QwQ respectively in subsequent text), QwQ-Plus and Deepseek-R1~\cite{guo2025deepseek}. A comprehensive overview of these models is presented in Table \ref{tab:models}.

All experiments were conducted using OpenCompass~\cite{opencompass}. For traditional instruction-based LLMs, we adopted OpenCompass's default configuration for maximum output length. For LRMs, we increased the output length to \(8192\) tokens to accommodate complex reasoning processes and prevent premature truncation.
For OpenAI models (GPT series and o1-mini), inference was performed via the official API, following OpenCompass's default settings. For Deepseek-V3 and Deepseek-R1, due to instability in Deepseek's official API under heavy usage, we used equivalent services provided by Alibaba Cloud (\url{https://cn.aliyun.com/}), maintaining OpenCompass's default parameters with a temperature of \(0.7\). All remaining models listed in Table~\ref{tab:models} were run locally on NVIDIA A100 GPUs, utilizing LMDeploy~\cite{lmdeploy} as the inference backend. In these cases, we used OpenCompass's standard settings, with \(\texttt{temperature} = 1\text{e-}6\) and \(\texttt{top\_k} = 1\).

\subsection{Evaluation Metrics}

Following \cite{wei2024measuring}, we use GPT-4o as a judge to classify the LLMs' responses into three categories: \textbf{\texttt{CORRECT}} (a response is correct if it fully contains the reference answer without contradictions), \textbf{\texttt{INCORRECT}} (applies when the predicted answer contradicts the reference answer), and \textbf{\texttt{NOT\_ATTEMPTED}} (if the response doesn't fully provide the reference answer but is free of contradictions). The judging prompt is detailed in Tables \ref{tab:llm_as_judge_1} and \ref{tab:llm_as_judge_2}.

Furthermore, we use five evaluation metrics to assess model performance.
\textbf{Correct (CO)}: The proportion of \texttt{CORRECT} answers among all answers.
\textbf{Not Attempted (NA)}: The proportion of \texttt{NOT\_ATTEMPTED} answers among all answers.
\textbf{Incorrect (IN)}: The proportion of \texttt{INCORRECT} answers among all answers.
\textbf{Correct Given Attempted (CGA)}: The proportion of \texttt{CORRECT} answers among those attempted.
\textbf{F-score}: A harmonic mean of CO and CGA, balancing correctness and successful attempts.
\section{Results \& Analysis}
\label{sec:Results}
\subsection{Overall Results and Domain Disparity}

The overall performance (F-score) of all LLMs on KolasSimpleQA is shown in Tables \ref{tab:overall_performance_fscore_1} and \ref{tab:overall_performance_fscore_2}, with additional metrics in Appendix \ref{sec:app_additional_res}. LLMs perform significantly worse in the specific domain than in the general one, highlighting challenges in language-specific factual QA. In the general domain, Deepseek-R1 and GPT4o achieved the top results, with only a 1.2 percentage point difference. However, in the language-specific domain, GPT4o led by nearly 7 percentage points, demonstrating its superior language-specific factual ability.

\begin{table}[htbp]
    \centering
    \begin{adjustbox}{width=\textwidth}
        \begin{tabular}{ccccccccccc}
    \toprule
         & \multicolumn{2}{c}{\textbf{zh}} & \multicolumn{2}{c}{\textbf{ko}} & \multicolumn{2}{c}{\textbf{th}} & \multicolumn{2}{c}{\textbf{ar}} & \multicolumn{2}{c}{\textbf{vi}} \\
\cmidrule{2-11}         & \textbf{general} & \textbf{specific} & \textbf{general} & \textbf{specific} & \textbf{general} & \textbf{specific} & \textbf{general} & \textbf{specific} & \textbf{general} & \textbf{specific} \\
    \midrule
    \textbf{GPT-4o-2024-11-20} & \textcolor[rgb]{ 0,  1,  0}{95.28} & \textcolor[rgb]{ 0,  0,  1}{23.95} & \textcolor[rgb]{ 0,  1,  0}{86.58} & \textcolor[rgb]{ 1,  0,  0}{38.36} & \textcolor[rgb]{ 0,  1,  0}{86.27} & \textcolor[rgb]{ 1,  0,  0}{31.28} & \textcolor[rgb]{ 0,  1,  0}{89.51} & \textcolor[rgb]{ 1,  0,  0}{38.89} & \textcolor[rgb]{ 1,  0,  0}{94.21} & \textcolor[rgb]{ 1,  0,  0}{54.55} \\
    \midrule
    \textbf{GPT-4o-mini} & 74.02 & 9.63 & 74.17 & \textcolor[rgb]{ 0,  0,  1}{28.92} & 67.63 & 16.49 & 72.84 & 19.30 & 80.17 & 34.97 \\
    \midrule
    \textbf{o1-mini} & 86.40 & 8.82 & 81.23 & 19.20 & 76.05 & 13.66 & 79.62 & 9.09 & 85.48 & 19.39 \\
    \midrule
    \textbf{Llama-3.1-Instruct-70B} & 86.51 & 11.76 & 78.15 & 22.89 & \textcolor[rgb]{ 0,  0,  1}{79.81} & 16.22 & 78.02 & 13.79 & 85.95 & 33.15 \\
    \midrule
    \textbf{Llama-3.1-Instruct-8B} & 45.85 & 5.41 & 42.38 & 13.58 & 52.40 & 11.40 & 48.75 & 3.79 & 69.42 & 20.11 \\
    \midrule
    \textbf{Qwen2.5-Instruct-72B} & 88.19 & 23.73 & 78.79 & 20.69 & 75.18 & 13.04 & 75.93 & 10.38 & 87.60 & 33.33 \\
    \midrule
    \textbf{Qwen2.5-Instruct-7B} & 57.71 & 14.57 & 39.59 & 7.94 & 45.15 & 8.28 & 34.89 & 6.67 & 60.00 & 19.88 \\
    \midrule
    \textbf{QwQ-32B} & 87.40 & 22.95 & 76.67 & 19.63 & 77.40 & 12.50 & 79.26 & 12.17 & 83.82 & 26.52 \\
    \midrule
    \textbf{QwQ-32B-Preview} & 79.03 & 19.05 & 74.56 & 17.60 & 72.40 & 13.85 & 78.29 & 9.88 & 81.03 & 20.13 \\
    \midrule
    \textbf{QwQ-Plus} & 63.32 & 15.48 & 57.85 & 19.85 & 50.32 & 10.39 & 60.70 & 9.63 & 61.38 & 18.18 \\
    \midrule
    \textbf{Deepseek\_V3} & \textcolor[rgb]{ 0,  0,  1}{94.49} & \textcolor[rgb]{ 0,  1,  0}{34.78} & \textcolor[rgb]{ 0,  0,  1}{82.12} & \textcolor[rgb]{ 0,  1,  0}{30.00} & 79.14 & \textcolor[rgb]{ 0,  1,  0}{19.59} & \textcolor[rgb]{ 0,  0,  1}{86.42} & \textcolor[rgb]{ 0,  0,  1}{22.81} & \textcolor[rgb]{ 0,  0,  1}{92.56} & \textcolor[rgb]{ 0,  0,  1}{44.94} \\
    \midrule
    \textbf{Deepseek\_R1} & \textcolor[rgb]{ 1,  0,  0}{96.85} & \textcolor[rgb]{ 1,  0,  0}{52.41} & \textcolor[rgb]{ 1,  0,  0}{86.75} & 25.30 & \textcolor[rgb]{ 1,  0,  0}{87.98} & \textcolor[rgb]{ 0,  1,  0}{19.59} & \textcolor[rgb]{ 1,  0,  0}{90.74} & \textcolor[rgb]{ 0,  1,  0}{25.00} & \textcolor[rgb]{ 0,  1,  0}{93.39} & \textcolor[rgb]{ 0,  1,  0}{52.17} \\
    \bottomrule
    \end{tabular}%
    \end{adjustbox}
    \caption{Model performance (F-score) on KolasSimpleQA (part1/2). \textcolor[rgb]{ 1,  0,  0}{Red} indicates the best, followed by \textcolor[rgb]{ 0,  1,  0}{green} and then \textcolor[rgb]{ 0,  0,  1}{blue}.}
  \label{tab:overall_performance_fscore_1}%
\end{table}%

\begin{table}[htbp]
    \centering
    \begin{adjustbox}{width=\textwidth}
        \begin{tabular}{ccccccccccc}
    \toprule
         & \multicolumn{2}{c}{\textbf{cs}} & \multicolumn{2}{c}{\textbf{hu}} & \multicolumn{2}{c}{\textbf{ru}} & \multicolumn{2}{c}{\textbf{sr}} & \multicolumn{2}{c}{\textbf{avg.}} \\
\cmidrule{2-11}         & \textbf{general} & \textbf{specific} & \textbf{general} & \textbf{specific} & \textbf{general} & \textbf{specific} & \textbf{general} & \textbf{specific} & \textbf{general} & \textbf{specific} \\
    \midrule
    \textbf{GPT-4o-2024-11-20} & \textcolor[rgb]{ 0,  0,  1}{91.87} & \textcolor[rgb]{ 1,  0,  0}{37.50} & \textcolor[rgb]{ 1,  0,  0}{88.17} & \textcolor[rgb]{ 1,  0,  0}{28.19} & \textcolor[rgb]{ 0,  0,  1}{86.34} & \textcolor[rgb]{ 1,  0,  0}{24.00} & \textcolor[rgb]{ 1,  0,  0}{93.58} & \textcolor[rgb]{ 1,  0,  0}{42.47} & \textcolor[rgb]{ 0,  1,  0}{90.20} & \textcolor[rgb]{ 1,  0,  0}{37.59} \\
    \midrule
    \textbf{GPT-4o-mini} & 83.80 & \textcolor[rgb]{ 0,  0,  1}{31.46} & 71.22 & 10.48 & 68.48 & 12.50 & 78.03 & 26.75 & 74.48 & 23.72 \\
    \midrule
    \textbf{o1-mini} & 84.89 & 24.83 & 73.27 & 10.58 & 79.78 & \textcolor[rgb]{ 0,  1,  0}{20.93} & 85.17 & 26.36 & 81.32 & 19.81 \\
    \midrule
    \textbf{Llama-3.1-Instruct-70B} & 89.44 & 26.82 & \textcolor[rgb]{ 0,  0,  1}{80.47} & \textcolor[rgb]{ 0,  0,  1}{16.87} & 83.06 & 15.93 & \textcolor[rgb]{ 0,  0,  1}{91.73} & \textcolor[rgb]{ 0,  1,  0}{31.37} & 83.68 & 23.56 \\
    \midrule
    \textbf{Llama-3.1-Instruct-8B} & 60.07 & 23.73 & 46.15 & 9.32 & 43.72 & 7.27 & 65.41 & 16.77 & 52.68 & 14.75 \\
    \midrule
    \textbf{Qwen2.5-Instruct-72B} & 81.69 & 21.30 & 52.38 & 6.31 & 72.13 & \textcolor[rgb]{ 0,  0,  1}{20.75} & 73.68 & 16.11 & 76.17 & 20.61 \\
    \midrule
    \textbf{Qwen2.5-Instruct-7B} & 50.18 & 8.70 & 35.11 & 0.00 & 32.58 & 6.32 & 33.59 & 15.60 & 43.20 & 12.11 \\
    \midrule
    \textbf{QwQ-32B} & 83.10 & 25.43 & 63.28 & 11.52 & 78.26 & 19.47 & 81.95 & 20.38 & 79.02 & 21.23 \\
    \midrule
    \textbf{QwQ-32B-Preview} & 79.41 & 16.79 & 65.18 & 9.47 & 79.77 & 18.67 & 78.43 & 24.14 & 76.46 & 18.34 \\
    \midrule
    \textbf{QwQ-Plus} & 70.00 & 10.37 & 48.69 & 5.49 & 60.00 & 14.43 & 53.06 & 13.91 & 58.37 & 14.70 \\
    \midrule
    \textbf{Deepseek\_V3} & \textcolor[rgb]{ 1,  0,  0}{92.96} & \textcolor[rgb]{ 0,  1,  0}{35.03} & 79.88 & 16.39 & \textcolor[rgb]{ 0,  1,  0}{87.43} & 16.36 & 88.72 & \textcolor[rgb]{ 0,  0,  1}{29.30} & \textcolor[rgb]{ 0,  0,  1}{87.08} & \textcolor[rgb]{ 0,  0,  1}{30.10} \\
    \midrule
    \textbf{Deepseek\_R1} & \textcolor[rgb]{ 0,  1,  0}{92.20} & 31.28 & \textcolor[rgb]{ 0,  1,  0}{87.83} & \textcolor[rgb]{ 0,  1,  0}{17.46} & \textcolor[rgb]{ 1,  0,  0}{92.39} & 14.16 & \textcolor[rgb]{ 1,  0,  0}{93.58} & 24.20 & \textcolor[rgb]{ 1,  0,  0}{91.30} & \textcolor[rgb]{ 0,  1,  0}{30.81} \\
    \bottomrule
    \end{tabular}%
    \end{adjustbox}
    \caption{Model performance (F-score) (part2/2). ``avg.'' denotes the average result across \textbf{all the 9} languages. \textcolor[rgb]{ 1,  0,  0}{Red} indicates the best, followed by \textcolor[rgb]{ 0,  1,  0}{green} and then \textcolor[rgb]{ 0,  0,  1}{blue}.}
  \label{tab:overall_performance_fscore_2}%
\end{table}%

\begin{figure}[t]
  \centering
  \includegraphics[width=1.0\linewidth]{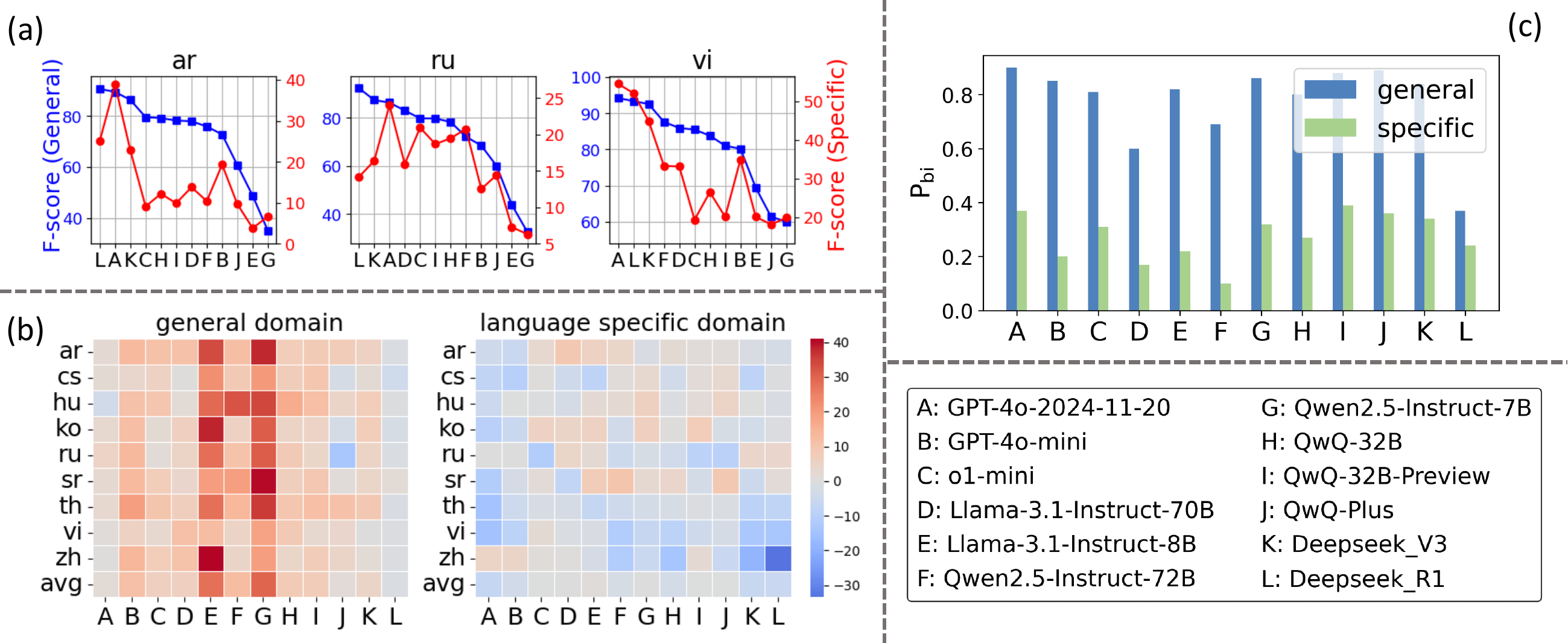}
  \caption{(a) Model performance (F-score) ranking in \textcolor{blue}{general} and \textcolor{red}{langugage-specific} domains. The models are sorted based on the general domain (\textcolor{blue}{blue} line).
  (b) Differences in F-scores between the   \texttt{tran\_en} and the \texttt{direct} settings (a value greater than zero indicates that \texttt{tran\_en} performs better).
  (c) Proportion of bidirectional correctness ($P_\text{bi}$) for general and specific domain questions across models.}
  \label{fig:rank_hotmap_bidirection}
\end{figure}

We compared model performance rankings between general and language-specific domains. Figure \ref{fig:rank_hotmap_bidirection}(a) and \ref{fig:appendix_model_rank} show that when models are ranked by general domain performance, their performance in language-specific domains shows significant fluctuations and jumps across nearly all 9 languages. This suggests that models optimized for the general domain may not excel in language-specific domains, emphasizing the need for targeted evaluation and optimization for language-specific tasks.

\subsection{Does Explicit translating into English help？ }

Research has shown \cite{shi2022language,huang2023not} that translating non-English questions into English before inputting them into an LLM significantly improves performance compared to using the original language directly. In our KoLasSimpleQA experiments, we established two settings: the \texttt{direct} setting,\footnote{Unless otherwise specified, we default to the \texttt{direct} setting.} where questions are input in their original language, and the \texttt{tran\_en} setting, where questions are first translated into English using GPT-4o\footnote{We use gpt-4o-08-06, which is not the model evaluated in this paper. See the prompt in Table~\ref{tab:prompt_translate_non_en_to_en}.} before being input into the LLM. Figure \ref{fig:rank_hotmap_bidirection}(b) details the performance differences between these settings across two domains. In the general domain, the \texttt{tran\_en} setting consistently enhances performance across most models and languages. However, in the language-specific domain, models generally perform better when questions are presented in their original language.

\subsection{Analysis of Calibration}

\begin{figure}[t]
  \centering
  \includegraphics[width=1.0\linewidth]{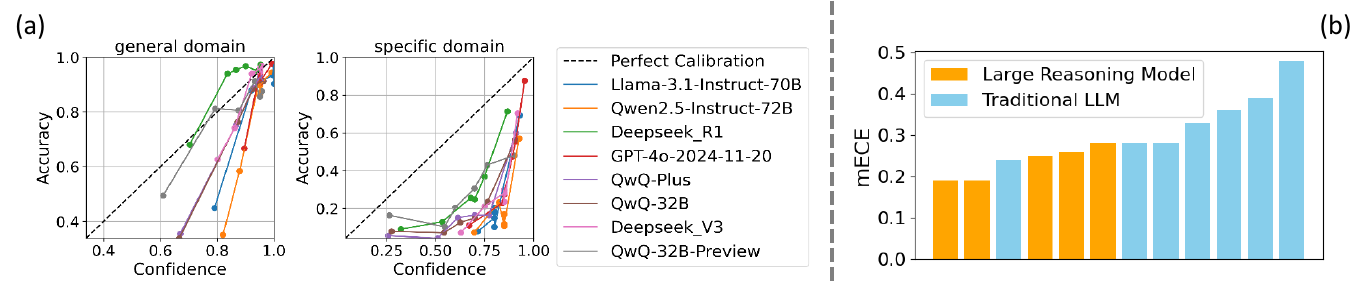}
  \caption{(a) Calibration of LLMs based on their expressed  confidence.
  (b) Mean of Expected Calibration Error (mECE), detailed results can be found in Table~\ref{tab:app_all_ece_result}.}
  \label{fig:calibration_curve}
\end{figure}

To evaluate LLMs' calibration, we prompted the LLM to provide a confidence level (0 to 100) when answering questions, as detailed in Table~\ref{tab:prompt_model_inference}. Figure \ref{fig:calibration_curve}(a) shows calibration performance across the two domains. We used the Expected Calibration Error (ECE) to quantitatively assess calibration. The ECE is calculated by dividing confidence scores into \(M\) equally spaced bins. For each bin \(B_m\), we compute the average confidence \(\text{conf}(B_m)\) and accuracy \(\text{acc}(B_m)\). ECE is defined as \(\sum_{m=1}^{M} \frac{|B_m|}{n} \left| \text{acc}(B_m) - \text{conf}(B_m) \right|\), where \(n\) is the total sample count, and \(|B_m|\) is the sample count in bin \(m\) (we set \(M = 10\)). Lower ECE indicates better calibration. We calculated ECE for each model across two domains, with results in Table~\ref{tab:app_all_ece_result}. To summarize, we averaged results across all 12 models to get the Average ECE (AvgECE), shown in Table~\ref{tab:AvgECE}. Additionally, we computed the mean ECE across both domains and settings for each model, called mECE, with results in Figure~\ref{fig:calibration_curve}(b) and \ref{fig:app_mECE}.

\setlength{\intextsep}{0.1em}
\setlength{\columnsep}{0.9em}
\begin{wraptable}{r}{0.45\textwidth}
\centering
\small
\vspace{0.1em}  
\begin{tabular}{@{}lll@{}}
\toprule
\textbf{}         & \textbf{general} & \textbf{specific} \\ \midrule
\textbf{\texttt{tran\_en}} & 0.06             & 0.48              \\
\textbf{\texttt{direct}}   & 0.13             & 0.5               \\ \bottomrule
\end{tabular}
\caption{Average Expected Calibration Error (AvgECE) across two domains and two settings (lower is better). Detailed results can be found in Table~\ref{tab:app_all_ece_result}.}
\label{tab:AvgECE}
\vspace{0.1em}  
\end{wraptable}

From these results, we conclude that:
(1) Models show significantly poorer calibration in the language-specific domain compared to the general domain.
(2) The \texttt{tran\_en} setting improves calibration in both domains, although it only enhances the F-score in the general domain. This indicates that LLMs not only vary in answering ability across languages but also in calibration, with English being the most effective language.
(3) LRMs demonstrate superior calibration compared to traditional LLMs, as all five LRMs rank within the top six for mECE values. This suggests that through thorough reasoning and reflection during inference, LRMs achieve better calibration performance than traditional LLMs.

\subsection{Knowledge Memorization Robustness}

Studies~\cite{allen2023physics,berglund2023reversal} have shown that auto-regressive LLMs struggle to generalize bidirectionally, a phenomenon known as the Reversal Curse. For instance, a model trained on "A's mother is B" may not correctly respond to "Who is B's child?" This reflects the robustness of knowledge retention in LLMs. In developing KoLasSimpleQA, we included QA pairs with reverse relationships, as illustrated in Figure~\ref{fig:Kolas examples reverse}, to evaluate the robustness of LLMs' knowledge memorization in multilingual contexts. We defined the metric Proportion of Bidirectional Correctness as 
\(
P_\text{bi} = {N_2}/{N_1}
\)
where \(N_2\) represents the number of reverse QA pairs where the LLM correctly answered at least one of the pair, and \(N_1\) is the number of pairs where the LLM answered both questions correctly. A higher \(P_\text{bi}\) indicates more robust memory of reverse knowledge points. As shown in Figure \ref{fig:rank_hotmap_bidirection}(c), models achieve significantly higher \(P_\text{bi}\) scores in the general domain compared to the language-specific domain. This suggests that general domain knowledge is more thoroughly represented in the pretraining data of LLMs, leading to more robust memorization, while language-specific domain knowledge is relatively scarce, underscoring the need for targeted optimization in these areas.

\subsection{Analyzing the Reasoning Process of LRMs}

\begin{figure}[htbp]
    \begin{minipage}[t]{0.45\textwidth}
        \centering
        \includegraphics[width=\textwidth]{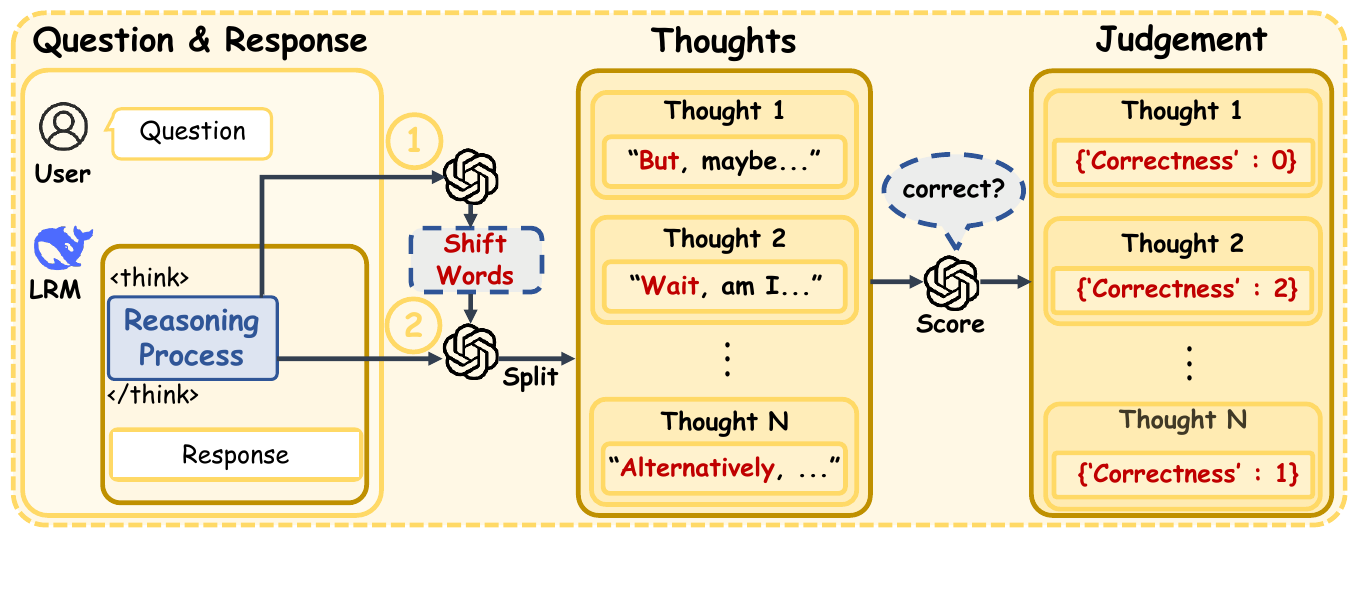}
        \captionof{figure}{Segment the LRM's reasoning process into distinct thoughts and judge the correctness of each.}
        \label{fig:LRM_reasoning_process_analysis_pipeline}
    \end{minipage}%
    \hfill
    \begin{minipage}[t]{0.48\textwidth}
        \centering
        \includegraphics[width=\textwidth]{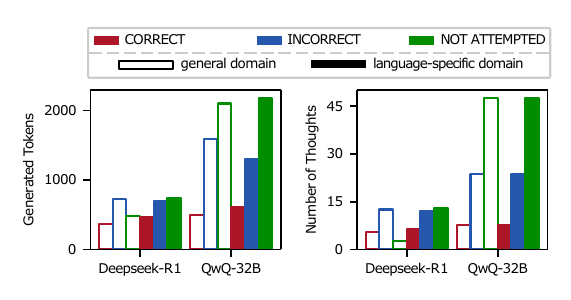}
        \captionof{figure}{Average number of tokens and thoughts generated during the LRM's reasoning process (\( \overline{n}_{\text{token}} \) and \( \overline{n}_{\text{thought}} \)).}
        \label{fig:LRM_tokens_thoughts_num}
    \end{minipage}
\end{figure}

Recently, LRMs, such as o1~\cite{openai-o1} and Deepseek\_R1\cite{guo2025deepseek}, represent a new direction in the development of LLMs. These models significantly enhance reasoning abilities through comprehensive reasoning, self-reflective negation, and exploration of multiple reasoning paths, adhering to the test-time scale law. However, research on their reasoning processes in multilingual scenarios remains underexplored. We address this by analyzing the reasoning processes of LRMs on KoLasSimpleQA, as shown in Figure~\ref{fig:LRM_reasoning_process_analysis_pipeline}.

When responding to user queries, an LRM's output typically includes two components: the \textbf{reasoning process} and the \textbf{answer}. As outlined in \cite{wang2025thoughts}, the reasoning process can be broken down into discrete "thoughts," which are intermediate steps generated during reasoning. The LRM transitions between these thoughts, often marked by reflective phrases like "Wait" or "Alternative." Examples are shown in Figures \ref{fig:app_ds_reasoning_example} and \ref{fig:app_qwq_reasoning_example}. We used GPT-4o to segment the reasoning process into distinct thoughts (see prompts in Tables \ref{tab:app_LRM_reasoning_analysis_prompt_shift_expression_extracting} and \ref{tab:app_LRM_reasoning_analysis_prompt_shift_expression_confirming}). Thoughts are classified as \textit{correct} or \textit{incorrect}: correct thoughts lead to \texttt{CORRECT} responses, while incorrect ones lead to \texttt{INCORRECT} responses. GPT-4o assessed the correctness of each thought (see Table \ref{tab:app_LRM_reasoning_analysis_prompt_thought_access}), with examples in Figures \ref{fig:app_ds_reasoning_example} and \ref{fig:app_qwq_reasoning_example}. The number of \textit{correct} thoughts in a response is denoted as \( n_{ct} \). A response with at least one \textit{correct} thought is labeled \( R_{n_{ct} \geq 1} \).

We applied the above process to responses in both the general and language-specific domains of KoLaSimpleQA. The LRM's responses were already categorized as \texttt{CORRECT}, \texttt{INCORRECT}, or \texttt{NOT\_ATTEMPTED} (refer to prompts in Tables \ref{tab:llm_as_judge_1} and \ref{tab:llm_as_judge_2}). We then conducted statistical analysis across these two domains and three categories, resulting in six distinct evaluations.

\begin{wrapfigure}{r}{0.5\textwidth}
    \centering
    \includegraphics[width=\linewidth]{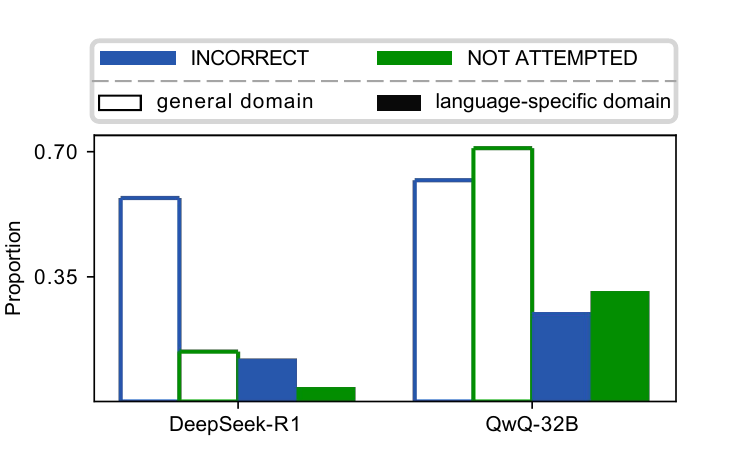}
    \caption{\( P(R_{n_{ct} \geq 1}) \) of the \texttt{INCORRECT} and \texttt{NOT\_ATTEMPTED} responses.}
    \label{fig:propotion_of_IN_NA_contains_correct_thought}
\end{wrapfigure}

To evaluate the LRM's reasoning process, we define the following metrics:
\textbf{(1) \( \overline{n}_{\text{token}} \) and \( \overline{n}_{\text{thought}} \):} These metrics represent the average number of tokens and thoughts generated during the reasoning process, respectively, providing a direct measure of its length and indicating the associated overhead.
\textbf{(2) \( P(R_{n_{ct} \geq 1}) \):} This metric denotes the proportion of \( R_{n_{ct} \geq 1} \) within a given category. We focus on this particularly in the \texttt{INCORRECT} and \texttt{NOT\_ATTEMPTED} categories, as it reflects instances where the LRM had the chance to answer correctly by "searching" through multiple thoughts but either answered incorrectly or chose not to attempt.

We selected two representative LRMs, Deepseek-R1 and QwQ-32B, and analyzed their reasoning processes on KoLasSimpleQA using the aforementioned methods. The results are shown in Figures \ref{fig:LRM_tokens_thoughts_num} and \ref{fig:propotion_of_IN_NA_contains_correct_thought}.
Based on these results, we found:
\textbf{(1) Overhead in the reasoning process}: As illustrated in Figure~\ref{fig:LRM_tokens_thoughts_num}, Deepseek-R1 maintains consistent reasoning costs across all three categories. Conversely, QwQ-32B shows significantly higher reasoning costs for \texttt{NOT\_ATTEMPTED} responses compared to the other categories. Notably, there is no significant difference in reasoning costs between the two domains for either model.
\textbf{(2) Regret in reasoning search}: Figure~\ref{fig:propotion_of_IN_NA_contains_correct_thought} shows that both models exhibit a much higher \( P(R_{n_{ct} \geq 1}) \) in the general domain than in the language-specific domain. This indicates that when tackling global knowledge questions, LRMs have substantial opportunities to answer correctly through thorough knowledge recall, search and reflection. However, due to various interferences and limitations in understanding their knowledge boundaries, they often miss the correct answer. In contrast, for language-specific knowledge questions, the lack of relevant pretraining knowledge becomes evident.
\section{Conclusion}
\label{sec:Conclusion}

This paper presents KoLasSimpleQA, a multilingual evaluation benchmark for assessing the factual capabilities of LLMs. KoLasSimpleQA focuses on two domains: general and language-specific. Comprehensive experiments show that in multilingual contexts, LLMs perform differently in language-specific versus general domains, highlighting the need for specialized evaluations and model optimization based on language specifics. We hope KoLasSimpleQA will advance LLM development and application in multilingual scenarios.

\section*{Acknowledgments}
This research was supported by National Key R\&D Program of China (2022ZD0160201).

\bibliography{colm2025_conference}
\bibliographystyle{colm2025_conference}

\appendix
\section{Related Work}
\label{sec:Related_Work}

\subsection{Multilingual Benchmark for General Domain}
The evaluation LLMs in the multilingual context has consistently been a focus of academic interest. A comprehensive summary of existing multilingual evaluation datasets can be found in this excellent review article (\cite{huang2024survey}). Many of these datasets are derived from translations of existing English evaluation sets. Earlier evaluation sets typically concentrated on individual tasks or capabilities, such as MGSM (\cite{shi2022language}), XNLI (\cite{conneau-etal-2018-xnli}), XCOPA (\cite{ponti-etal-2020-xcopa}), and BELEBELE (\cite{bandarkar2023belebele}). Recent studies, such as PMMEval (\cite{zhang2024p}) and BenchMaX (\cite{huang2025benchmax}), utilize translated parallel corpora to comprehensively assess the performance of LLMs across multiple tasks and capability dimensions in multilingual settings.
These evaluation sets, based on translated parallel corpora, allow for a thorough assessment of the language-agnostic abilities of LLMs, bu overlook the unique characteristics and capabilities inherent to most of the world's languages, such as culture, history, geography, religion, and local life, which are crucial and highly relevant to the speakers of these languages.

\subsection{Multilingual Benchmark for language-specific domain}
There are various methods for constructing benchmarks that evaluate language-specific features (\cite{liu2024culturally}). Some works collect exam questions from different countries to build benchmarks similar to MMLU (\cite{hendrycks2020measuring}), such as CMMLU (\cite{li-etal-2024-cmmlu}), CEval (\cite{huang2023c}), IndoMMLU (\cite{koto-etal-2023-large}), ArabicMMLU (\cite{koto2024arabicmmlu}), KMMLU (\cite{son2024kmmlu}), TurkishMMLU (\cite{yuksel2024turkishmmlu}), and MMLU-ProX (\cite{xuan2025mmlu}). However, their data is limited in scale, and most of the knowledge remains global, such as STEM knowledge. 
Some involves manually constructing evaluation sets (\cite{myung2024blend, sun2024benchmarking, ligeti-nagy-etal-2024-hulu}), but this method also faces scalability challenges. 
Another approach involves crawling content from internet forums and user queries, followed by filtering to retain only those queries relevant to language features, to build evaluation sets (\cite{naous2023having, arora2024calmqa}). 
Moreover, cultural-related materials (such as concepts, proverbs, etc.) are collected as seeds, and evaluation sets are constructed through an "LLM generation + human expert revision/inspection" approach (\cite{liu2023multilingual, putri-etal-2024-llm}), which is a recent trend.
However, the problems in the aforementioned evaluation sets are typically complex: first, they include both factual inquiries and reasoning abilities; second, they usually involve queries about multiple facts or knowledge points; third, the answers are often open-ended, with no absolute or unique standard answer. This paper, following the framework of SimpleQA, constructs a multilingual version that effectively encompasses both global and language-specific domains, thereby facilitating the evaluation of LLM's factual abilities in multilingual contexts through "simple" questions.

\subsection{Dataset based on Wikipedia}
Wikipedia plays a crucial role in the development of large language models (LLMs). It serves not only as the corpus for nearly all LLM pretraining (\cite{chen2020mining}), but also as the foundation for constructing SFT data (\cite{wang-etal-2024-ultralink}) and evaluation datasets (\cite{yu2023kola, he2024chinese, he2024mintqa}). In particular, Wikipedia contains entries in XX languages, which are typically composed of native speaker-generated content across various languages (\cite{5365214, yang2010motivations}), making it a valuable repository of native multilingual data.
However, the Wikipedia corpus encompasses vast amounts of global knowledge, which, despite being presented in non-English forms (e.g., STEM-related entries, world knowledge), may not fully capture the linguistic characteristics of each language. This paper leverages the meta-information in Wikipedia pages to effectively distinguish entries, filtering out those that genuinely contain language-specific knowledge, such as history, geography, people, and events. Based on this, we have constructed a benchmark dataset that authentically reflects linguistic features.

\section{Prompt Templates Used in KolasSimpleQA Construction and Qualify Control}
\label{sec:app_constuction}

We show all the prompt templates used in the construction of KolasSimpleQA. The prompt template for extracting the triples is shown in Table~\ref{tab:prompt_extract_triple}. The prompt template for constructing the QA pair from the triples is shown in Table~\ref{tab:prompt_construct_qa_from_triple}.
The prompt template for translating the original non-English question of KolasSimpleQA into English is shown in Table~\ref{tab:prompt_translate_non_en_to_en}.
The prompt template for quality control (stage1) is shown in Table~\ref{tab:prompt_quality_control_stage1}.
The prompt template for quality control (stage2) is shown in Table~\ref{tab:llm_as_judge_3}.

\begin{table}[t]
\centering

\begin{tcolorbox}[title={Extract Triple Prompt}, colback=white, coltitle=black, colbacktitle=white!0]

You are a professional natural language processing assistant, responsible for extracting structured relational triples from text in the format of \texttt{[[`Entity A', `Relation 1', `Entity B'], [`Entity B', `Relation 2', `Entity A']]}. Please extract all the triples that meet the requirements from the following text, where:
\\ 
1. Entity A and Entity B are entities explicitly mentioned in the text.\\
2. Relation is the core meaning of the verb, phrase or sentence that describes the association between entity A and entity B.\\

The output triples are represented in the form of a list, and ensure that:

1. The triples are accurate and based on the text content.\\
2. Do not contain subjective inferences, and only extract clear textual relations.\\
3. If the corresponding relational triples between the two entities cannot be extracted based on the text, output \texttt{[]}.\\
4. The language of relations and entities of the extracted triples are consistent with the language of the provided text.\\
5. For Entity A and Entity B, it is necessary to extract the relationship triples between Entity A and Entity B and the relationship triples between Entity B and Entity A at the same time. The final output format is: \texttt{[[`Entity A', `Relation 1', `Entity B'], [`Entity B', `Relation 2', `Entity A']]}.

Output format: \\
\texttt{[[`Entity A', `Relation 1', `Entity B'], [`Entity B', `Relation 2', `Entity A']]} \\
Here is am example:

\textbf{[Input text]:} \\
Li Sing Primary School Li Sing Primary School (English: Li Sing Primary School) is a government primary school located in Sai Ying Pun, Hong Kong. It was founded in 1954. In May 1953, Li Baochun announced that he would invest 250,000 yuan to open this primary school. The school site is the former site of Sai Ying Pun Government School. ...

\textbf{[Entity A]:} Li Baochun \\
\textbf{[Entity B]:} Li Sing Primary School

\textbf{[Output Result]:}
\begin{verbatim}
{
    "triple_pair": [['Li Baochun', 'Opened', 'Li Sing Primary School'],['Li 
    Sheng Primary School', 'is Opened by', 'Li Baochun']]
}
\end{verbatim}

\textbf{[Input Text]:} \texttt{<context>} \\
\textbf{[Entity A]:} \texttt{<entity1>} \\
\textbf{[Entity B]:} \texttt{<entity2>}

\textbf{[Output Result]:}
\begin{verbatim}
{
    "triple_pair": []
}
\end{verbatim}

\end{tcolorbox}
\caption{Prompt template for triple extraction in the construction of KoLasSimpleQA.}
\label{tab:prompt_extract_triple}
\end{table}

\begin{table}[t]
\centering
\small
\begin{tcolorbox}[title={Construct Question-Answer Prompt}, colback=white, coltitle=black, colbacktitle=white!0]

As a general knowledge expert, please generate open-ended questions that can be answered independently based on the specified knowledge material and triples related to the material, and ensure that the questions meet the following requirements:

1. Given a piece of material and the triples extracted from the material, generate questions based on the triples. Each question is an independent question that can be answered independently without the materials. The question can contain appropriate context materials for simple background explanation (such as the attributive that describes the entity, specify the time when the event occurred, etc.) to avoid ambiguity.

2. The question stem must specify the scope of the answer. For example, instead of asking "where did Barack and Michelle Obama meet" (for which the answers could be "Chicago" or "the law firm Sidley \& Austin"), the question should specify "which city" or "which company". Another example: instead of asking "when", ask "which year" or "which day".

3. Reference answers should not change over time. For example, instead of broadly asking "who is Meredith's partner in Grey's Anatomy", which could change as new seasons are produced, questions asking about TV shows, movies, video games, and sports typically require specifying a point in time (e.g., "who is Meredith's partner in Grey's Anatomy in Season 13").

4. Questions must have a clear and unique answer (the tail entity in the triple). For example, for the triple ["China", "contains", "Beijing"], the question cannot be "Which province does China contain?" because the answer is not unique. For triples that cannot generate questions, output \texttt{None}. Another example: do not ask "What is one of the representative pieces of the Mei'an Qin School?" as the answer is not unique.

5. A triple and its corresponding opposite triple will be provided. If the provided triple is \texttt{[]}, the corresponding question and answer should be \texttt{None}.

6. The question is about the head entity of the triple, and the answer is the tail entity of the triple.

7. The given \texttt{[triple]} is used to generate \texttt{[question]} and \texttt{[answer]}, and the given \texttt{[reverse triple]} is used to generate \texttt{[question\_reverse]} and \texttt{[answer\_reverse]}. The head entity and the relation of the triple are used to generate the question, and the tail entity is the answer.

8. Language: The questions are in \texttt{<language>}.
\\
Here is an example:

\textbf{[Input materials]:} 
Li Sing Primary School

Li Sing Primary School (English: Li Sing Primary School) is a government primary school located in Sai Ying Pun, Hong Kong. It was founded in 1954. 
In May 1953, Li Baochun announced that he would invest 250,000 yuan to open this primary school. The school site is the former site of Sai Ying Pun Government School. ...

\textbf{[triple]:} \texttt{['Li Baochun', 'Opened', 'Li Sing Primary School']} \\
\textbf{[reverse triple]:} \texttt{['Li Sing Primary School', 'is Opened by', 'Li Baochun']}

\textbf{[Output Result]:}
\begin{verbatim}
{
    "question": "What is the name of the primary school opened by Li Baochun?",
    "answer": "i Sheng Primary School",
    "question_reverse": "Who is the founder of Li Sheng Primary School?",
    "answer_reverse": "Li Baochun"
}
\end{verbatim}

\textbf{[Input materials]:} \texttt{<context>} \\
\textbf{[triple]:} \texttt{<triple>} \\
\textbf{[reverse triple]:} \texttt{<reverse\_triple>}

\textbf{[Output Result]:}
\begin{verbatim}
{
    "question": "",
    "answer": "",
    "question_reverse": "",
    "answer_reverse": ""
}
\end{verbatim}

\end{tcolorbox}
\caption{Prompt template for constructing questions from triples in the construction of KoLasSimpleQA.}
\label{tab:prompt_construct_qa_from_triple}
\end{table}

\begin{table}[t]
\centering

\begin{tcolorbox}[title={Translatation Prompt}, colback=white, coltitle=black, colbacktitle=white!0]

You are a language expert specialized in \textit{from\_lang} and \textit{to\_lang}. Please translate the following open-ended question and its answer into \textit{to\_lang}. 
Ensure that the semantics and format are consistent with those before translation. Try to translate names of people and places into the target language.

\textbf{[Question]}\\
\texttt{<question>}

\textbf{[Answer]}\\
\texttt{<answer>}

\textbf{[Output Result]:}\\
Please respond strictly in JSON format. Do not include any additional text outside the JSON structure:
\begin{verbatim}
{
    "question_trans": [the translation of question],
    "answer_trans":[the translation of answer]
}  
\end{verbatim}

\end{tcolorbox}
\caption{Prompt template for translating the original non-English question of KoLasSimpleQA into English}
\label{tab:prompt_translate_non_en_to_en}
\end{table}

\begin{table}[t]
\centering
\fontsize{8pt}{9.6pt}\selectfont

\begin{tcolorbox}[title={Quality Control Prompt}, colback=white, coltitle=black, colbacktitle=white!0]

You are a knowledge question quality inspection expert. Your task is to evaluate the quality of knowledge test questions generated from given materials and extracted triples. Each triple is in the format \texttt{['head entity', 'relation', 'tail entity']}. Please assess the quality based on the following criteria:

1. The extracted triples must be correct and consistent with the input materials.\\
2. Questions must be generated using the head entity and relation of the triple; the answer must be the tail entity.\\
3. Questions must contain all necessary context and be answerable independently without access to the original material.\\
4. Questions should not be overly simple; the answer must not be directly revealed in the question stem.\\
5. Questions must target objective knowledge and yield a single, indisputable answer. For instance, do not ask "Where did Barack and Michelle Obama meet?" (which could have multiple answers like "Chicago" or "Sidley Austin LLP"). Instead, specify "which city" or "which company". Similarly, avoid vague time expressions like "when" and use precise ones like "which year" or "which date".\\
6. Questions must have time-invariant answers. Avoid asking questions whose answers change over time. For example, do not ask "Who is Meredith's partner on Grey's Anatomy?" Instead, specify the season, e.g., "Who is Meredith's partner in Season 13?"\\
7. If the triple fails the quality check, then the corresponding question must also be judged as failing.\\
8. Questions must have a clear and unique answer (i.e., the tail entity). For instance, for the triple \texttt{["China", "contains", "Beijing"]}, the question "Which province does China contain?" is invalid, as it has multiple possible answers. Similarly, avoid vague questions like "What is one of the representative pieces of the Mei'an Qin School?"\\
9. The language used in the question must be \texttt{<language>}.

\vspace{0.3em}
\textbf{Output format:} 
\begin{verbatim}
{
    "check_triple": "[whether the triple is correct]", 
    "check_independent": "[whether the question can be answered independently 
    without input material]", 
    "check_answer_is_tail": "[whether the answer is the tail entity of triple]", 
    "check_unique": "[whether the answer to the question is unique]", 
    "check_question": "[whether the question and answer is correct]", 
    "check_question_reason": "[the reason why the question and answer is true or false]"
}
\end{verbatim}
Here is an example

\textbf{[Input materials]:} 
Ibn al-Kizani was born in Egypt and lived in the sixth century AH.

\textbf{[triple]:} \texttt{['Egypt', 'is birthplace', 'Ibn al-Kizani']} \\
\textbf{[question]:} Who was born in Egypt and lived in the sixth century AH? \\
\textbf{[answer]:} Ibn al-Kizani

\textbf{[Output Result]:}
\begin{verbatim}
{
    "check_triple": true, 
    "check_independent": false, 
    "check_answer_is_tail": true, 
    "check_unique": false, 
    "check_question": false, 
    "check_question_reason": "The answer is not unique and the question is not 
    independently answerable. It fails."
}
\end{verbatim}

Please strictly follow the format below to generate your output:

\textbf{[Input materials]:} \texttt{<context>} \\
\textbf{[triple]:} \texttt{<triple>} \\
\textbf{[question]:} \texttt{<question>} \\
\textbf{[answer]:} \texttt{<answer>}

\textbf{[Output Result]:}
\begin{verbatim}
{
    "check_triple": true/false,
    "check_independent": true/false,
    "check_answer_is_tail": true/false,
    "check_unique": true/false,
    "check_question": true/false,
    "check_question_reason": "[your explanation]"
}
\end{verbatim}

\end{tcolorbox}
\caption{Prompt template for qualify control (stage 1) in the construction of KoLasSimpleQA.}
\label{tab:prompt_quality_control_stage1}
\end{table}

\begin{table}[t]
\centering
\small
\begin{tcolorbox}[title={Judge Prompt}, colback=white, coltitle=black, colbacktitle=white!0]

Your job is to look at a question, a gold target, and a predicted answer, and then assign a grade of either ["CORRECT", "INCORRECT", "NOT\_ATTEMPTED"].

\textbf{Question:} \{question\}\\
\textbf{Gold target:} \{target\}\\
\textbf{Predicted answer:} \{predicted\_answer\}\\

Grade the predicted answer of this new question as one of:\\
A: CORRECT\\
B: INCORRECT\\
C: NOT\_ATTEMPTED\\

Just return the letters "A", "B", or "C", with no text around it.

\end{tcolorbox}
\caption{Prompt template for quality control(stage 2) in the construction of KoLasSimpleQA.}
\label{tab:llm_as_judge_3}
\end{table}

\section{Details of Evaluations}
\label{sec:app_infer_judge}

The prompt template for model inference is shown in Table~\ref{tab:prompt_model_inference}.
The prompt template for LLM as judge is shown in Table~\ref{tab:llm_as_judge_1} and~\ref{tab:llm_as_judge_2}.

\begin{table}[t]
\centering

\begin{tcolorbox}[title={Infer Prompt}, colback=white, coltitle=black, colbacktitle=white!0]

Here is the question: \\
\{question\} \\
Please provide your best guess and a confidence score between 0 to 100 in the following JSON format: 

\begin{verbatim}
{ 
  "answer": "Your answer here", 
  "confidence_score": number 
} 
\end{verbatim}

\end{tcolorbox}
\caption{Prompt template for model inference.}
\label{tab:prompt_model_inference}
\end{table}

\begin{table}[t]
\centering
\small
\begin{tcolorbox}[title={Judge Prompt}, colback=white, coltitle=black, colbacktitle=white!0]

Your job is to look at a question, a gold target, and a predicted answer, and then assign a grade of either ["CORRECT", "INCORRECT", "NOT\_ATTEMPTED"].\\
First, I will give examples of each grade, and then you will grade a new example.\\

The following are examples of CORRECT predicted answers:\\

\textbf{Question:} What are the names of Barack Obama's children?\\
\textbf{Gold target:} Malia Obama and Sasha Obama\\
\textbf{Predicted answer 1:} sasha and malia obama.\\
\textbf{Predicted answer 2:} most people would say Malia and Sasha, but I'm not sure and would have to double check.\\
\textbf{Predicted answer 3:} Barack Obama has two daughters. Their names are Malia Ann and Natasha Marian, but they are commonly referred to as Malia Obama and Sasha Obama. Malia was born on July 4, 1998, and Sasha was born on June 10, 2001.\\

These predicted answers are all CORRECT because:
\begin{verbatim}
  - They fully contain the important information in the gold target.
  - They do not contain any information that contradicts the gold target.
  - Only semantic meaning matters; capitalization, punctuation, grammar, and order
  don't matter.
  - Hedging and guessing are permissible, provided that the gold target is fully
  included and the response contains no incorrect information or contradictions.
\end{verbatim}

The following are examples of INCORRECT predicted answers:\\

\textbf{Question:} What are the names of Barack Obama's children?\\
\textbf{Gold target:} Malia and Sasha\\
\textbf{Predicted answer 1:} Malia.\\
\textbf{Predicted answer 2:} Malia, Sasha, and Susan.\\
\textbf{Predicted answer 3:} Barack Obama does not have any children.\\
\textbf{Predicted answer 4:} I think it's either Malia and Sasha. Or it could be Malia and Jackie. Or it could be Joey and Malia.\\
\textbf{Predicted answer 5:} While I don't know their exact names, I can tell you that Barack Obama has three children.\\
\textbf{Predicted answer 6:} It's possible you may mean Betsy and Olivia. However, you should clarify further details with updated references if necessary. Is that the correct answer?\\
\textbf{Predicted answer 7:} It may be the case that Obama's child is named James. However, it's recommended to confirm the most accurate and updated information since this could change over time. This model may not always reflect the most current information.\\

These predicted answers are all INCORRECT because:
\begin{verbatim}
    - A factual statement in the answer contradicts the gold target. Incorrect statements
    that have some hedging (e.g., "it is possible that", "although i'm not sure, i think "
    ) are also considered incorrect.
\end{verbatim}

The following are examples of NOT\_ATTEMPTED predicted answers:\\

\textbf{Question:} What are the names of Barack Obama's children?\\
\textbf{Gold target:} Malia and Sasha\\
\textbf{Predicted answer 1:} I don't know.\\
\textbf{Predicted answer 2:} I need more context about which Obama you are talking about.\\
\textbf{Predicted answer 3:} Without researching the web, I cannot answer this question. However, I can tell you that Barack Obama has two children.\\
\textbf{Predicted answer 4:} Barack Obama has two children. I know that one of them is Malia, but I'm not sure about the other one.\\

These predicted answers are all NOT\_ATTEMPTED because:
\begin{verbatim}
  - The important information in the gold target is not included in the answer.
  - No statements in the answer contradict the gold target.
\end{verbatim}

\end{tcolorbox}
\caption{Prompt template for LLM as judge (part1/2).}
\label{tab:llm_as_judge_1}
\end{table}
\begin{table}[t]
\centering
\small
\begin{tcolorbox}[title={Judge Prompt}, colback=white, coltitle=black, colbacktitle=white!0]

Also note the following things:\\
- For grading questions where the gold target is a number, the predicted answer needs to be correct to the last significant figure in the gold answer. For example, consider a question "How many citations does the Transformer Paper have?" with gold target "120k".
\begin{verbatim}
  - Predicted answers "120000", "120k" are all CORRECT.
  - Predicted answers "100k" and "113k" are INCORRECT.
  - Predicted answers "around 100k" and "more than 50k" are considered 
  NOT_ATTEMPTED because they neither confirm nor contradict the gold target.
\end{verbatim}
  
- The gold target may contain more information than the question. In such cases, the predicted answer only needs to contain the information that is in the question.
\begin{verbatim}
  - For example, consider the question "What episode did Derek and Meredith get
  legally married in Grey's Anatomy?" with gold target "Season 7, Episode 20:
  White Wedding". Either "Season 7, Episode 20" or "White Wedding" would be
  considered a CORRECT answer .
\end{verbatim}
  
- Do not punish predicted answers if they omit information that would be clearly inferred from the question.
\begin{verbatim}
  - For example, consider the question "What city is OpenAI headquartered in?"
  and the gold target "San Francisco, California". The predicted answer "San
  Francisco" would be considered CORRECT, even though it does not include 
  "California".
  - Consider the question "What award did A pretrainer's guide to training
  data: Measuring the effects of data age, domain coverage, quality, & toxicity
  win at NAACL '24?", the gold target is "Outstanding Paper Award". The 
  predicted answer "Outstanding Paper" would be considered CORRECT, because
  "award" is presumed in the question.
  - For the question "What is the height of Jason Wei in meters?", the gold
  target is "1.73 m". The predicted answer "1.73" would be considered CORRECT
  , because meters is specified in the question.
  - For the question "What is the name of Barack Obama's wife?", the gold
  target is " Michelle Obama". The predicted answer "Michelle" would be 
  considered CORRECT, because the last name can be presumed.
\end{verbatim}
  
- Do not punish for typos in people's name if it's clearly the same name.
\begin{verbatim}
  - For example, if the gold target is "Hyung Won Chung", you can consider
  the following predicted answers as correct: "Hyoong Won Choong", 
  "Hyungwon Chung", or "Hyun Won Chung".    
\end{verbatim}

- Do not punish if the language type of predicted answer is different from that of question.
Here is a new example. Simply reply with either CORRECT, INCORRECT, NOT ATTEMPTED. Don't apologize or correct yourself if there was a mistake; we are just trying to grade the answer.\\

\textbf{Question:} \{question\}\\
\textbf{Gold target:} \{target\}\\
\textbf{Predicted answer:} \{predicted\_answer\}\\

Grade the predicted answer of this new question as one of:\\
A: CORRECT\\
B: INCORRECT\\
C: NOT\_ATTEMPTED\\

Just return the letters "A", "B", or "C", with no text around it.

\end{tcolorbox}
\caption{Prompt template for LLM as judge (part2/2).}
\label{tab:llm_as_judge_2}
\end{table}
\section{Additional Results}
\label{sec:app_additional_res}

The details of the evaluated LLMs are listed in Table~\ref{tab:models}.

The detailed model performance is outlined below:
\begin{itemize}
    \item Correct (CO): See Tables~\ref{tab:res_CO_1} and~\ref{tab:res_CO_2}.
    \item Not Attempted (NA): See Tables~\ref{tab:res_NA_1} and~\ref{tab:res_NA_2}.
    \item Incorrect (IN): See Tables~\ref{tab:res_IN_1} and~\ref{tab:res_IN_2}.
    \item Correct Given Attempted (CGA): See Tables~\ref{tab:res_CGA_1} and~\ref{tab:res_CGA_2}.
\end{itemize}

\begin{figure}[t]
  \centering
  \includegraphics[width=1.0\linewidth]{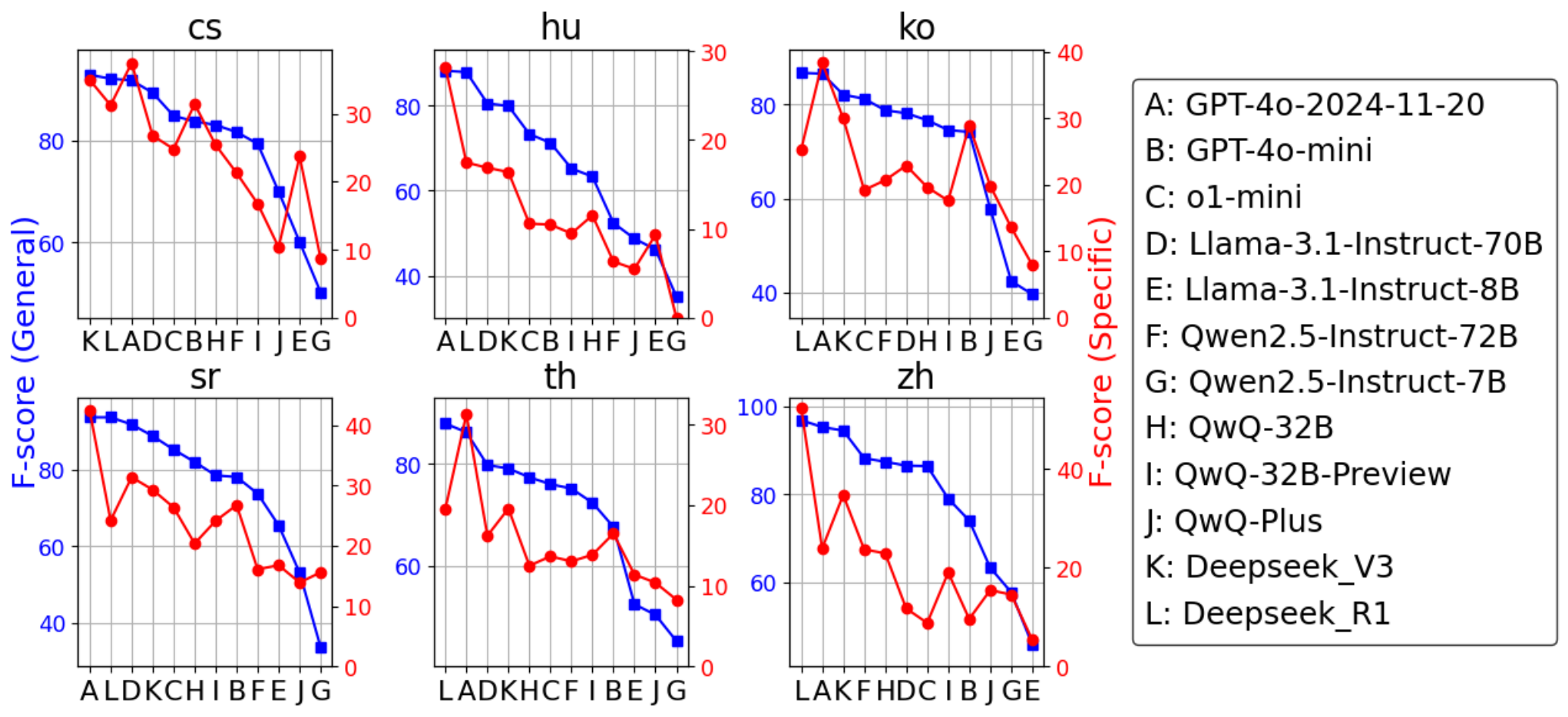}
  \caption{Model performance (F-score) ranking in \textcolor{blue}{general} and \textcolor{red}{langugage-specific} domains. The models are sorted based on the general domain (\textcolor{blue}{blue} line).}
  \label{fig:appendix_model_rank}
\end{figure}

The model performance (F-score) ranking in the general and the langugage-specific domains are shown in Figure~\ref{fig:rank_hotmap_bidirection}(a) and~\ref{fig:appendix_model_rank}.

The detail of the Expected Calibration Error (ECE) is shown in Table~\ref{tab:app_all_ece_result}.

\begin{figure}[t]
  \centering
  \includegraphics[width=0.8\linewidth]{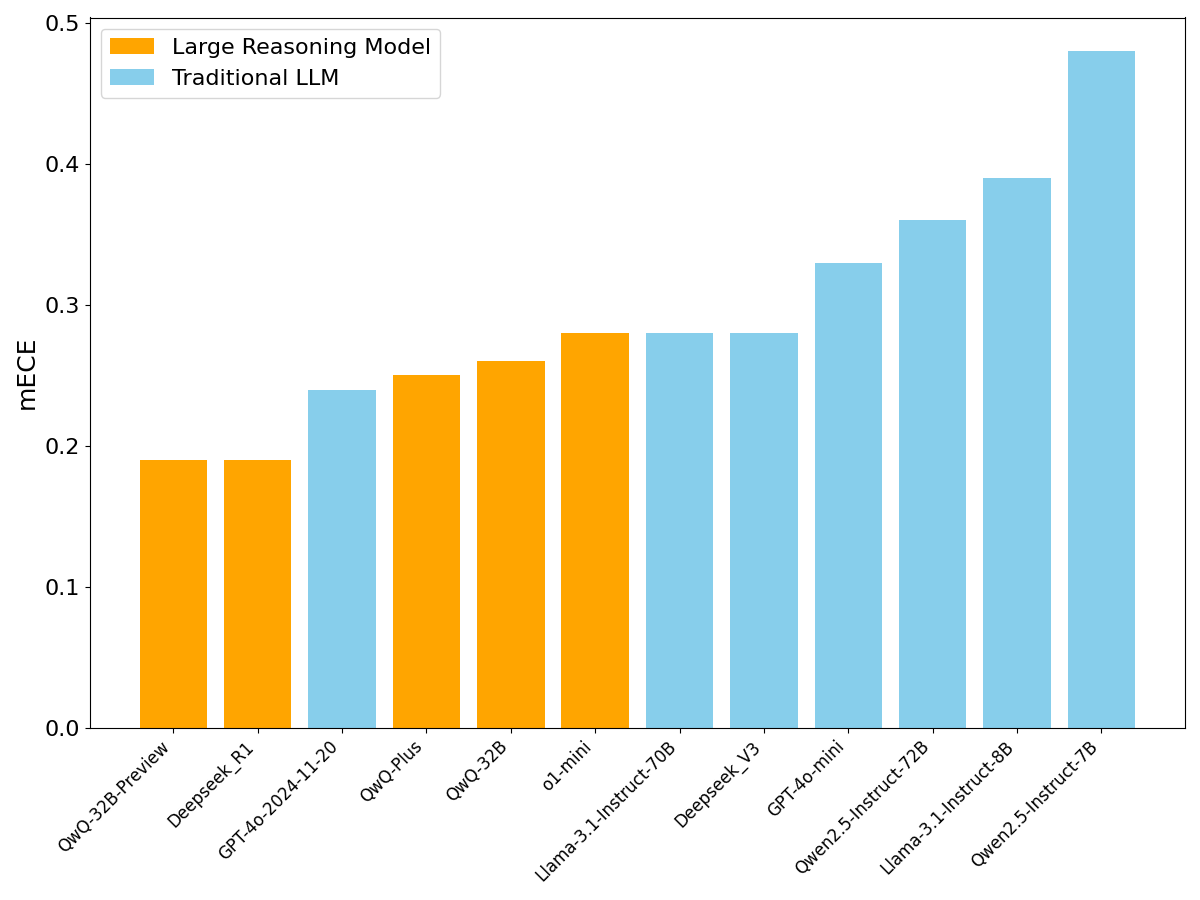}
  \caption{Mean of Expected Calibration Error (mECE), detailed results can be found in Table~\ref{tab:app_all_ece_result}.}
  \label{fig:app_mECE}
\end{figure}

The Mean of Expected Calibration Error (mECE) is shown in Figure~\ref{fig:app_mECE}.

\begin{table}[t]
\centering
\begin{tabular}{ccccc}
\textbf{Model} &
\textbf{Size} &
\begin{tabular}[c]{@{}c@{}}\textbf{Reasoning}\\ \textbf{Model}\end{tabular} &
\begin{tabular}[c]{@{}c@{}}\textbf{Open-}\\ \textbf{source}\end{tabular} &
\textbf{Inference Method} \\ \hline
\rowcolor[HTML]{EFEFEF} 
\cellcolor[HTML]{EFEFEF} &
  \cellcolor[HTML]{EFEFEF} &
  \cellcolor[HTML]{EFEFEF} &
  \cellcolor[HTML]{EFEFEF} &
  \cellcolor[HTML]{EFEFEF} \\
\rowcolor[HTML]{EFEFEF} 
\multirow{-2}{*}{\cellcolor[HTML]{EFEFEF}GPT-4o} &
  \multirow{-2}{*}{\cellcolor[HTML]{EFEFEF}-} &
  \multirow{-2}{*}{\cellcolor[HTML]{EFEFEF}N} &
  \multirow{-2}{*}{\cellcolor[HTML]{EFEFEF}N} &
  \multirow{-2}{*}{\cellcolor[HTML]{EFEFEF}Official API} \\
GPT-4o-mini &-
   &
  N &
  N &
  Official API \\
\rowcolor[HTML]{EFEFEF} 
\cellcolor[HTML]{EFEFEF} &
  \cellcolor[HTML]{EFEFEF} &
  \cellcolor[HTML]{EFEFEF} &
  \cellcolor[HTML]{EFEFEF} &
  \cellcolor[HTML]{EFEFEF} \\
\rowcolor[HTML]{EFEFEF} 
\multirow{-2}{*}{\cellcolor[HTML]{EFEFEF}Deepseek-V3} &
  \multirow{-2}{*}{\cellcolor[HTML]{EFEFEF}-} &
  \multirow{-2}{*}{\cellcolor[HTML]{EFEFEF}N} &
  \multirow{-2}{*}{\cellcolor[HTML]{EFEFEF}Y} &
  \multirow{-2}{*}{\cellcolor[HTML]{EFEFEF}\begin{tabular}[c]{@{}c@{}}Alibaba Cloud \end{tabular}} \\
Qwen2.5-Instruct &
  7B,72B &
  N &
  Y &
  Local GPU \\
\rowcolor[HTML]{EFEFEF} 
Llama-3.1-Instruct &
  8B,70B &
  N &
  Y &
  Local GPU \\
o1-mini &
  - &
  Y &
  N &
  Official API \\
\rowcolor[HTML]{EFEFEF} 
QwQ &
  32B &
  Y &
  Y &
  Local GPU \\
  QwQ-preview &
  32B &
  Y &
  Y &
  Local GPU \\
  \rowcolor[HTML]{EFEFEF} 
QwQ-Plus &
  - &
  Y &
  N &
  Official API \\
  Deepseek-R1 &
  - &
  Y &
  N &
  Official API \\
\end{tabular}
\caption{LLMs evaluated in our experiments. }
\label{tab:models}
\end{table}

\begin{table}[htbp]
    \centering
    \begin{adjustbox}{width=\textwidth}
        \begin{tabular}{ccccccccccc}
    \toprule
         & \multicolumn{2}{c}{\textbf{zh}} & \multicolumn{2}{c}{\textbf{ko}} & \multicolumn{2}{c}{\textbf{th}} & \multicolumn{2}{c}{\textbf{ar}} & \multicolumn{2}{c}{\textbf{vi}} \\
\cmidrule{2-11}         & \textbf{general} & \textbf{specific} & \textbf{general} & \textbf{specific} & \textbf{general} & \textbf{specific} & \textbf{general} & \textbf{specific} & \textbf{general} & \textbf{specific} \\
    \midrule
    \textbf{GPT-4o-2024-11-20} & \textcolor[rgb]{ 0,  1,  0}{95.28} & 21.05 & \textcolor[rgb]{ 0,  1,  0}{85.43} & \textcolor[rgb]{ 1,  0,  0}{33.33} & \textcolor[rgb]{ 0,  1,  0}{85.65} & \textcolor[rgb]{ 1,  0,  0}{28.28} & \textcolor[rgb]{ 0,  1,  0}{89.51} & \textcolor[rgb]{ 1,  0,  0}{35.90} & \textcolor[rgb]{ 1,  0,  0}{94.21} & \textcolor[rgb]{ 1,  0,  0}{52.17} \\
    \midrule
    \textbf{GPT-4o-mini} & 74.02 & 9.47 & 74.17 & \textcolor[rgb]{ 0,  1,  0}{28.57} & 67.46 & 16.16 & 72.84 & 18.80 & 80.17 & 34.78 \\
    \midrule
    \textbf{o1-mini} & 85.04 & 6.32 & 78.81 & 14.29 & 73.68 & 11.11 & 78.40 & 6.84 & 85.12 & 17.39 \\
    \midrule
    \textbf{Llama-3.1-Instruct-70B} & 85.83 & 11.58 & 78.15 & 22.62 & \textcolor[rgb]{ 0,  0,  1}{79.43} & 15.15 & 77.78 & 13.68 & 85.95 & 32.61 \\
    \midrule
    \textbf{Llama-3.1-Instruct-8B} & 45.67 & 5.26 & 42.38 & 13.10 & 52.15 & 11.11 & 48.15 & 3.42 & 69.42 & 19.57 \\
    \midrule
    \textbf{Qwen2.5-Instruct-72B} & 88.19 & \textcolor[rgb]{ 0,  0,  1}{22.11} & 77.48 & 17.86 & 73.21 & 12.12 & 75.93 & 9.40 & 87.60 & 31.52 \\
    \midrule
    \textbf{Qwen2.5-Instruct-7B} & 57.48 & 11.58 & 38.41 & 5.95 & 44.50 & 7.07 & 34.57 & 5.98 & 59.50 & 17.39 \\
    \midrule
    \textbf{QwQ-32B} & 87.40 & \textcolor[rgb]{ 0,  0,  1}{22.11} & 76.16 & 19.05 & 77.03 & 12.12 & 79.01 & 11.97 & 83.47 & 26.09 \\
    \midrule
    \textbf{QwQ-32B-Preview} & 77.17 & 14.74 & 70.86 & 13.10 & 66.51 & 9.09 & 73.46 & 6.84 & 77.69 & 16.30 \\
    \midrule
    \textbf{QwQ-Plus} & 49.61 & 12.63 & 46.36 & 15.48 & 37.80 & 8.08 & 48.15 & 7.69 & 47.93 & 14.13 \\
    \midrule
    \textbf{Deepseek\_V3} & \textcolor[rgb]{ 0,  0,  1}{94.49} & \textcolor[rgb]{ 0,  1,  0}{33.68} & \textcolor[rgb]{ 0,  0,  1}{82.12} & \textcolor[rgb]{ 0,  1,  0}{28.57} & 78.95 & \textcolor[rgb]{ 0,  1,  0}{19.19} & \textcolor[rgb]{ 0,  0,  1}{86.42} & \textcolor[rgb]{ 0,  0,  1}{22.22} & \textcolor[rgb]{ 0,  0,  1}{92.56} & \textcolor[rgb]{ 0,  0,  1}{43.48} \\
    \midrule
    \textbf{Deepseek\_R1} & \textcolor[rgb]{ 1,  0,  0}{96.85} & \textcolor[rgb]{ 1,  0,  0}{51.58} & \textcolor[rgb]{ 1,  0,  0}{86.75} & 25.00 & \textcolor[rgb]{ 1,  0,  0}{87.56} & \textcolor[rgb]{ 0,  1,  0}{19.19} & \textcolor[rgb]{ 1,  0,  0}{90.74} & \textcolor[rgb]{ 0,  1,  0}{24.79} & \textcolor[rgb]{ 0,  1,  0}{93.39} & \textcolor[rgb]{ 1,  0,  0}{52.17} \\
    \bottomrule
    \end{tabular}%
    \end{adjustbox}
    \caption{Model performance (CO) on KoLasSimpleQA (part1/2).}
  \label{tab:res_CO_1}%
\end{table}%

\begin{table}[htbp]
    \centering
    \begin{adjustbox}{width=\textwidth}
        \begin{tabular}{ccccccccccc}
    \toprule
         & \multicolumn{2}{c}{\textbf{cs}} & \multicolumn{2}{c}{\textbf{hu}} & \multicolumn{2}{c}{\textbf{ru}} & \multicolumn{2}{c}{\textbf{sr}} & \multicolumn{2}{c}{\textbf{avg.}} \\
\cmidrule{2-11}         & \textbf{general} & \textbf{specific} & \textbf{general} & \textbf{specific} & \textbf{general} & \textbf{specific} & \textbf{general} & \textbf{specific} & \textbf{general} & \textbf{specific} \\
    \midrule
    \textbf{GPT-4o-2024-11-20} & \textcolor[rgb]{ 0,  1,  0}{91.55} & \textcolor[rgb]{ 0,  1,  0}{33.33} & \textcolor[rgb]{ 1,  0,  0}{88.17} & \textcolor[rgb]{ 1,  0,  0}{25.20} & \textcolor[rgb]{ 0,  0,  1}{85.87} & \textcolor[rgb]{ 1,  0,  0}{21.05} & \textcolor[rgb]{ 1,  0,  0}{93.23} & \textcolor[rgb]{ 1,  0,  0}{39.24} & \textcolor[rgb]{ 0,  1,  0}{89.88} & \textcolor[rgb]{ 1,  0,  0}{34.15} \\
    \midrule
    \textbf{GPT-4o-mini} & 83.80 & \textcolor[rgb]{ 0,  0,  1}{31.11} & 71.01 & 10.24 & 68.48 & 12.28 & 77.44 & 26.58 & 74.38 & 23.38 \\
    \midrule
    \textbf{o1-mini} & 83.10 & 20.00 & 72.19 & 7.87 & 77.17 & 15.79 & 84.21 & 21.52 & 79.75 & 15.69 \\
    \midrule
    \textbf{Llama-3.1-Instruct-70B} & 89.44 & 26.67 & \textcolor[rgb]{ 0,  0,  1}{80.47} & \textcolor[rgb]{ 0,  0,  1}{16.54} & 82.61 & 15.79 & \textcolor[rgb]{ 0,  0,  1}{91.73} & \textcolor[rgb]{ 0,  1,  0}{30.38} & 83.49 & 23.13 \\
    \midrule
    \textbf{Llama-3.1-Instruct-8B} & 59.86 & 23.33 & 46.15 & 8.66 & 43.48 & 7.02 & 65.41 & 16.46 & 52.52 & 14.34 \\
    \midrule
    \textbf{Qwen2.5-Instruct-72B} & 81.69 & 20.00 & 52.07 & 5.51 & 71.74 & \textcolor[rgb]{ 0,  1,  0}{19.30} & 73.68 & 15.19 & 75.73 & 19.02 \\
    \midrule
    \textbf{Qwen2.5-Instruct-7B} & 48.59 & 7.78 & 33.14 & 0.00 & 31.52 & 5.26 & 33.08 & 13.92 & 42.31 & 10.28 \\
    \midrule
    \textbf{QwQ-32B} & 83.10 & 24.44 & 62.72 & 11.02 & 78.26 & \textcolor[rgb]{ 0,  1,  0}{19.30} & 81.95 & 20.25 & 78.79 & 20.66 \\
    \midrule
    \textbf{QwQ-32B-Preview} & 76.06 & 12.22 & 60.36 & 6.30 & 75.00 & 12.28 & 75.19 & 17.72 & 72.48 & 13.36 \\
    \midrule
    \textbf{QwQ-Plus} & 59.15 & 7.78 & 38.46 & 3.94 & 48.91 & 12.28 & 39.10 & 10.13 & 46.16 & 11.46 \\
    \midrule
    \textbf{Deepseek\_V3} & \textcolor[rgb]{ 1,  0,  0}{92.96} & \textcolor[rgb]{ 1,  0,  0}{34.44} & 79.88 & 15.75 & \textcolor[rgb]{ 0,  1,  0}{86.96} & 15.79 & 88.72 & \textcolor[rgb]{ 0,  0,  1}{29.11} & \textcolor[rgb]{ 0,  0,  1}{87.01} & \textcolor[rgb]{ 0,  0,  1}{29.19} \\
    \midrule
    \textbf{Deepseek\_R1} & \textcolor[rgb]{ 0,  1,  0}{91.55} & \textcolor[rgb]{ 0,  0,  1}{31.11} & \textcolor[rgb]{ 0,  1,  0}{87.57} & \textcolor[rgb]{ 0,  1,  0}{17.32} & \textcolor[rgb]{ 1,  0,  0}{92.39} & 14.04 & \textcolor[rgb]{ 1,  0,  0}{93.23} & 24.05 & \textcolor[rgb]{ 1,  0,  0}{91.11} & \textcolor[rgb]{ 0,  1,  0}{30.48} \\
    \bottomrule
    \end{tabular}%
    \end{adjustbox}
    \caption{Model performance (CO) on KoLasSimpleQA (part2/2).}
  \label{tab:res_CO_2}%
\end{table}%

\begin{table}[htbp]
    \centering
    \begin{adjustbox}{width=\textwidth}
    \begin{tabular}{ccccccccccc}
    \toprule
         & \multicolumn{2}{c}{\textbf{zh}} & \multicolumn{2}{c}{\textbf{ko}} & \multicolumn{2}{c}{\textbf{th}} & \multicolumn{2}{c}{\textbf{ar}} & \multicolumn{2}{c}{\textbf{vi}} \\
\cmidrule{2-11}         & \textbf{general} & \textbf{specific} & \textbf{general} & \textbf{specific} & \textbf{general} & \textbf{specific} & \textbf{general} & \textbf{specific} & \textbf{general} & \textbf{specific} \\
    \midrule
    \textbf{GPT-4o-2024-11-20} & 0.00 & 24.21 & 2.65 & 26.19 & 1.44 & 19.19 & 0.00 & 15.38 & 0.00 & 8.70 \\
    \midrule
    \textbf{GPT-4o-mini} & 0.00 & 3.16 & 0.00 & 2.38 & 0.48 & 4.04 & 0.00 & 5.13 & 0.00 & 1.09 \\
    \midrule
    \textbf{o1-mini} & \textcolor[rgb]{ 0,  0,  1}{3.15} & \textcolor[rgb]{ 1,  0,  0}{56.84} & \textcolor[rgb]{ 0,  0,  1}{5.96} & \textcolor[rgb]{ 1,  0,  0}{51.19} & \textcolor[rgb]{ 0,  0,  1}{6.22} & \textcolor[rgb]{ 0,  0,  1}{37.37} & \textcolor[rgb]{ 0,  0,  1}{3.09} & \textcolor[rgb]{ 0,  1,  0}{49.57} & 0.83 & 20.65 \\
    \midrule
    \textbf{Llama-3.1-Instruct-70B} & 1.57 & 3.16 & 0.00 & 2.38 & 0.96 & 13.13 & 0.62 & 1.71 & 0.00 & 3.26 \\
    \midrule
    \textbf{Llama-3.1-Instruct-8B} & 0.79 & 5.26 & 0.00 & 7.14 & 0.96 & 5.05 & 2.47 & 19.66 & 0.00 & 5.43 \\
    \midrule
    \textbf{Qwen2.5-Instruct-72B} & 0.00 & 13.68 & 3.31 & 27.38 & 5.26 & 14.14 & 0.00 & 18.80 & 0.00 & 10.87 \\
    \midrule
    \textbf{Qwen2.5-Instruct-7B} & 0.79 & \textcolor[rgb]{ 0,  1,  0}{41.05} & \textcolor[rgb]{ 0,  0,  1}{5.96} & \textcolor[rgb]{ 0,  0,  1}{50.00} & 2.87 & 29.29 & 1.85 & 20.51 & \textcolor[rgb]{ 0,  0,  1}{1.65} & \textcolor[rgb]{ 0,  0,  1}{25.00} \\
    \midrule
    \textbf{QwQ-32B} & 0.00 & 7.37 & 1.32 & 5.95 & 0.96 & 6.06 & 0.62 & 3.42 & 0.83 & 3.26 \\
    \midrule
    \textbf{QwQ-32B-Preview} & \textcolor[rgb]{ 0,  1,  0}{4.72} & \textcolor[rgb]{ 0,  0,  1}{45.26} & \textcolor[rgb]{ 0,  1,  0}{9.93} & \textcolor[rgb]{ 1,  0,  0}{51.19} & \textcolor[rgb]{ 0,  1,  0}{16.27} & \textcolor[rgb]{ 1,  0,  0}{68.69} & \textcolor[rgb]{ 0,  1,  0}{12.35} & \textcolor[rgb]{ 1,  0,  0}{61.54} & \textcolor[rgb]{ 0,  1,  0}{8.26} & \textcolor[rgb]{ 0,  1,  0}{38.04} \\
    \midrule
    \textbf{QwQ-Plus} & \textcolor[rgb]{ 1,  0,  0}{43.31} & 36.84 & \textcolor[rgb]{ 1,  0,  0}{39.74} & 44.05 & \textcolor[rgb]{ 1,  0,  0}{49.76} & \textcolor[rgb]{ 0,  1,  0}{44.44} & \textcolor[rgb]{ 1,  0,  0}{41.36} & \textcolor[rgb]{ 0,  0,  1}{40.17} & \textcolor[rgb]{ 1,  0,  0}{43.80} & \textcolor[rgb]{ 1,  0,  0}{44.57} \\
    \midrule
    \textbf{Deepseek\_V3} & 0.00 & 6.32 & 0.00 & 9.52 & 0.48 & 4.04 & 0.00 & 5.13 & 0.00 & 6.52 \\
    \midrule
    \textbf{Deepseek\_R1} & 0.00 & 3.16 & 0.00 & 2.38 & 0.96 & 4.04 & 0.00 & 1.71 & 0.00 & 0.00 \\
    \bottomrule
    \end{tabular}%
    \end{adjustbox}
    \caption{Model performance (NA) on KoLasSimpleQA (part1/2).}
  \label{tab:res_NA_1}%
\end{table}%

\begin{table}[htbp]
    \centering
    \begin{adjustbox}{width=\textwidth}
        \begin{tabular}{ccccccccccc}
    \toprule
         & \multicolumn{2}{c}{\textbf{cs}} & \multicolumn{2}{c}{\textbf{hu}} & \multicolumn{2}{c}{\textbf{ru}} & \multicolumn{2}{c}{\textbf{sr}} & \multicolumn{2}{c}{\textbf{avg.}} \\
\cmidrule{2-11}         & \textbf{general} & \textbf{specific} & \textbf{general} & \textbf{specific} & \textbf{general} & \textbf{specific} & \textbf{general} & \textbf{specific} & \textbf{general} & \textbf{specific} \\
    \midrule
    \textbf{GPT-4o-2024-11-20} & 0.70 & 22.22 & 0.00 & 21.26 & 1.09 & 24.56 & 0.75 & 15.19 & 0.74 & 19.13 \\
    \midrule
    \textbf{GPT-4o-mini} & 0.00 & 2.22 & 0.59 & 4.72 & 0.00 & 3.51 & 1.50 & 1.27 & 0.29 & 3.17 \\
    \midrule
    \textbf{o1-mini} & 4.23 & \textcolor[rgb]{ 0,  0,  1}{38.89} & 2.96 & \textcolor[rgb]{ 0,  0,  1}{51.18} & \textcolor[rgb]{ 0,  0,  1}{6.52} & \textcolor[rgb]{ 0,  1,  0}{49.12} & 2.26 & \textcolor[rgb]{ 0,  0,  1}{36.71} & 3.91 & \textcolor[rgb]{ 0,  0,  1}{42.28} \\
    \midrule
    \textbf{Llama-3.1-Instruct-70B} & 0.00 & 1.11 & 0.00 & 3.94 & 1.09 & 1.75 & 0.00 & 6.33 & 0.47 & 3.89 \\
    \midrule
    \textbf{Llama-3.1-Instruct-8B} & 0.70 & 3.33 & 0.00 & 14.17 & 1.09 & 7.02 & 0.00 & 3.80 & 0.67 & 7.60 \\
    \midrule
    \textbf{Qwen2.5-Instruct-72B} & 0.00 & 12.22 & 1.18 & 25.20 & 1.09 & 14.04 & 0.00 & 11.39 & 1.20 & 16.59 \\
    \midrule
    \textbf{Qwen2.5-Instruct-7B} & \textcolor[rgb]{ 0,  0,  1}{6.34} & 21.11 & \textcolor[rgb]{ 0,  0,  1}{11.24} & 48.03 & \textcolor[rgb]{ 0,  0,  1}{6.52} & \textcolor[rgb]{ 0,  0,  1}{33.33} & \textcolor[rgb]{ 0,  0,  1}{3.01} & 21.52 & \textcolor[rgb]{ 0,  0,  1}{4.47} & 31.66 \\
    \midrule
    \textbf{QwQ-32B} & 0.00 & 7.78 & 1.78 & 8.66 & 0.00 & 1.75 & 0.00 & 1.27 & 0.61 & 5.16 \\
    \midrule
    \textbf{QwQ-32B-Preview} & \textcolor[rgb]{ 0,  1,  0}{8.45} & \textcolor[rgb]{ 1,  0,  0}{54.44} & \textcolor[rgb]{ 0,  1,  0}{14.79} & \textcolor[rgb]{ 1,  0,  0}{66.93} & \textcolor[rgb]{ 0,  1,  0}{11.96} & \textcolor[rgb]{ 1,  0,  0}{68.42} & \textcolor[rgb]{ 0,  1,  0}{8.27} & \textcolor[rgb]{ 0,  1,  0}{53.16} & \textcolor[rgb]{ 0,  1,  0}{10.56} & \textcolor[rgb]{ 1,  0,  0}{56.63} \\
    \midrule
    \textbf{QwQ-Plus} & \textcolor[rgb]{ 1,  0,  0}{30.99} & \textcolor[rgb]{ 0,  1,  0}{50.00} & \textcolor[rgb]{ 1,  0,  0}{42.01} & \textcolor[rgb]{ 0,  1,  0}{56.69} & \textcolor[rgb]{ 1,  0,  0}{36.96} & 29.82 & \textcolor[rgb]{ 1,  0,  0}{52.63} & \textcolor[rgb]{ 1,  0,  0}{54.43} & \textcolor[rgb]{ 1,  0,  0}{42.28} & \textcolor[rgb]{ 0,  1,  0}{45.25} \\
    \midrule
    \textbf{Deepseek\_V3} & 0.00 & 3.33 & 0.00 & 7.87 & 1.09 & 7.02 & 0.00 & 1.27 & 0.17 & 5.89 \\
    \midrule
    \textbf{Deepseek\_R1} & 1.41 & 1.11 & 0.59 & 1.57 & 0.00 & 1.75 & 0.75 & 1.27 & 0.41 & 2.00 \\
    \bottomrule
    \end{tabular}%
    \end{adjustbox}
    \caption{Model performance (NA) on KoLasSimpleQA (part2/2).}
  \label{tab:res_NA_2}%
\end{table}%

\begin{table}[htbp]
    \centering
    \begin{adjustbox}{width=\textwidth}
        \begin{tabular}{ccccccccccc}
    \toprule
         & \multicolumn{2}{c}{\textbf{zh}} & \multicolumn{2}{c}{\textbf{ko}} & \multicolumn{2}{c}{\textbf{th}} & \multicolumn{2}{c}{\textbf{ar}} & \multicolumn{2}{c}{\textbf{vi}} \\
\cmidrule{2-11}         & \textbf{general} & \textbf{specific} & \textbf{general} & \textbf{specific} & \textbf{general} & \textbf{specific} & \textbf{general} & \textbf{specific} & \textbf{general} & \textbf{specific} \\
    \midrule
    \textbf{GPT-4o-2024-11-20} & 4.72 & 54.74 & 11.92 & 40.48 & 12.92 & 52.53 & 10.49 & 48.72 & 5.79 & 39.13 \\
    \midrule
    \textbf{GPT-4o-mini} & \textcolor[rgb]{ 0,  0,  1}{25.98} & \textcolor[rgb]{ 0,  1,  0}{87.37} & \textcolor[rgb]{ 0,  0,  1}{25.83} & 69.05 & \textcolor[rgb]{ 0,  0,  1}{32.06} & \textcolor[rgb]{ 0,  0,  1}{79.80} & \textcolor[rgb]{ 0,  0,  1}{27.16} & 76.07 & \textcolor[rgb]{ 0,  0,  1}{19.83} & \textcolor[rgb]{ 0,  0,  1}{64.13} \\
    \midrule
    \textbf{o1-mini} & 11.81 & 36.84 & 15.23 & 34.52 & 20.10 & 51.52 & 18.52 & 43.59 & 14.05 & 61.96 \\
    \midrule
    \textbf{Llama-3.1-Instruct-70B} & 12.60 & \textcolor[rgb]{ 0,  0,  1}{85.26} & 21.85 & \textcolor[rgb]{ 0,  1,  0}{75.00} & 19.62 & 71.72 & 21.60 & \textcolor[rgb]{ 1,  0,  0}{84.62} & 14.05 & \textcolor[rgb]{ 0,  0,  1}{64.13} \\
    \midrule
    \textbf{Llama-3.1-Instruct-8B} & \textcolor[rgb]{ 1,  0,  0}{53.54} & \textcolor[rgb]{ 1,  0,  0}{89.47} & \textcolor[rgb]{ 1,  0,  0}{57.62} & \textcolor[rgb]{ 1,  0,  0}{79.76} & \textcolor[rgb]{ 0,  1,  0}{46.89} & \textcolor[rgb]{ 1,  0,  0}{83.84} & \textcolor[rgb]{ 0,  1,  0}{49.38} & \textcolor[rgb]{ 0,  0,  1}{76.92} & \textcolor[rgb]{ 0,  1,  0}{30.58} & \textcolor[rgb]{ 1,  0,  0}{75.00} \\
    \midrule
    \textbf{Qwen2.5-Instruct-72B} & 11.81 & 64.21 & 19.21 & 54.76 & 21.53 & 73.74 & 24.07 & 71.79 & 12.40 & 57.61 \\
    \midrule
    \textbf{Qwen2.5-Instruct-7B} & \textcolor[rgb]{ 0,  1,  0}{41.73} & 47.37 & \textcolor[rgb]{ 0,  1,  0}{55.63} & 44.05 & \textcolor[rgb]{ 1,  0,  0}{52.63} & 63.64 & \textcolor[rgb]{ 1,  0,  0}{63.58} & 73.50 & \textcolor[rgb]{ 1,  0,  0}{38.84} & 57.61 \\
    \midrule
    \textbf{QwQ-32B} & 12.60 & 70.53 & 22.52 & \textcolor[rgb]{ 0,  1,  0}{75.00} & 22.01 & \textcolor[rgb]{ 0,  1,  0}{81.82} & 20.37 & \textcolor[rgb]{ 1,  0,  0}{84.62} & 15.70 & \textcolor[rgb]{ 0,  1,  0}{70.65} \\
    \midrule
    \textbf{QwQ-32B-Preview} & 18.11 & 40.00 & 19.21 & 35.71 & 17.22 & 22.22 & 14.20 & 31.62 & 14.05 & 45.65 \\
    \midrule
    \textbf{QwQ-Plus} & 7.09 & 50.53 & 13.91 & 40.48 & 12.44 & 47.47 & 10.49 & 52.14 & 8.26 & 41.30 \\
    \midrule
    \textbf{Deepseek\_V3} & 5.51 & 60.00 & 17.88 & 61.90 & 20.57 & 76.77 & 13.58 & 72.65 & 7.44 & 50.00 \\
    \midrule
    \textbf{Deepseek\_R1} & 3.15 & 45.26 & 13.25 & 72.62 & 11.48 & 76.77 & 9.26 & 73.50 & 6.61 & 47.83 \\
    \bottomrule
    \end{tabular}%
    \end{adjustbox}
    \caption{Model performance (IN) on KoLasSimpleQA (part1/2).}
  \label{tab:res_IN_1}%
\end{table}%

\begin{table}[htbp]
    \centering
    \begin{adjustbox}{width=\textwidth}
    \begin{tabular}{ccccccccccc}
    \toprule
         & \multicolumn{2}{c}{\textbf{cs}} & \multicolumn{2}{c}{\textbf{hu}} & \multicolumn{2}{c}{\textbf{ru}} & \multicolumn{2}{c}{\textbf{sr}} & \multicolumn{2}{c}{\textbf{avg.}} \\
\cmidrule{2-11}         & \textbf{general} & \textbf{specific} & \textbf{general} & \textbf{specific} & \textbf{general} & \textbf{specific} & \textbf{general} & \textbf{specific} & \textbf{general} & \textbf{specific} \\
    \midrule
    \textbf{GPT-4o-2024-11-20} & 7.75 & 44.44 & 11.83 & 53.54 & 13.04 & 54.39 & 6.02 & 45.57 & 9.39 & 46.72 \\
    \midrule
    \textbf{GPT-4o-mini} & 16.20 & 66.67 & 28.40 & \textcolor[rgb]{ 1,  0,  0}{85.04} & \textcolor[rgb]{ 0,  0,  1}{31.52} & \textcolor[rgb]{ 0,  1,  0}{84.21} & 21.05 & 72.15 & \textcolor[rgb]{ 0,  0,  1}{25.34} & \textcolor[rgb]{ 0,  0,  1}{73.45} \\
    \midrule
    \textbf{o1-mini} & 12.68 & 41.11 & 24.85 & 40.94 & 16.30 & 35.09 & 13.53 & 41.77 & 16.34 & 42.03 \\
    \midrule
    \textbf{Llama-3.1-Instruct-70B} & 10.56 & \textcolor[rgb]{ 0,  1,  0}{72.22} & 19.53 & 79.53 & 16.30 & 82.46 & 8.27 & 63.29 & 16.04 & 72.98 \\
    \midrule
    \textbf{Llama-3.1-Instruct-8B} & \textcolor[rgb]{ 0,  1,  0}{39.44} & \textcolor[rgb]{ 1,  0,  0}{73.33} & \textcolor[rgb]{ 0,  1,  0}{53.85} & 77.17 & \textcolor[rgb]{ 0,  1,  0}{55.43} & \textcolor[rgb]{ 1,  0,  0}{85.96} & \textcolor[rgb]{ 0,  1,  0}{34.59} & \textcolor[rgb]{ 1,  0,  0}{79.75} & \textcolor[rgb]{ 0,  1,  0}{46.81} & \textcolor[rgb]{ 1,  0,  0}{78.06} \\
    \midrule
    \textbf{Qwen2.5-Instruct-72B} & \textcolor[rgb]{ 0,  0,  1}{18.31} & 67.78 & \textcolor[rgb]{ 0,  0,  1}{46.75} & 69.29 & 27.17 & 66.67 & \textcolor[rgb]{ 0,  0,  1}{26.32} & 73.42 & 23.06 & 64.39 \\
    \midrule
    \textbf{Qwen2.5-Instruct-7B} & \textcolor[rgb]{ 1,  0,  0}{45.07} & \textcolor[rgb]{ 0,  0,  1}{71.11} & \textcolor[rgb]{ 1,  0,  0}{55.62} & 51.97 & \textcolor[rgb]{ 1,  0,  0}{61.96} & 61.40 & \textcolor[rgb]{ 1,  0,  0}{63.91} & 64.56 & \textcolor[rgb]{ 1,  0,  0}{53.22} & 58.06 \\
    \midrule
    \textbf{QwQ-32B} & 16.90 & 67.78 & 35.50 & \textcolor[rgb]{ 0,  0,  1}{80.31} & 21.74 & 78.95 & 18.05 & \textcolor[rgb]{ 0,  1,  0}{78.48} & 20.60 & \textcolor[rgb]{ 0,  1,  0}{74.18} \\
    \midrule
    \textbf{QwQ-32B-Preview} & 15.49 & 33.33 & 24.85 & 26.77 & 13.04 & 19.30 & 16.54 & 29.11 & 16.97 & 30.01 \\
    \midrule
    \textbf{QwQ-Plus} & 9.86 & 42.22 & 19.53 & 39.37 & 14.13 & 57.89 & 8.27 & 35.44 & 11.55 & 43.29 \\
    \midrule
    \textbf{Deepseek\_V3} & 7.04 & 62.22 & 20.12 & 76.38 & 11.96 & 77.19 & 11.28 & 69.62 & 12.82 & 64.91 \\
    \midrule
    \textbf{Deepseek\_R1} & 7.04 & 67.78 & 11.83 & \textcolor[rgb]{ 0,  1,  0}{81.10} & 7.61 & \textcolor[rgb]{ 0,  1,  0}{84.21} & 6.02 & \textcolor[rgb]{ 0,  0,  1}{74.68} & 8.47 & 67.52 \\
    \bottomrule
    \end{tabular}%
    \end{adjustbox}
    \caption{Model performance (IN) on KoLasSimpleQA (part2/2).}
  \label{tab:res_IN_2}%
\end{table}%

\begin{table}[htbp]
    \centering
    \begin{adjustbox}{width=\textwidth}
        \begin{tabular}{ccccccccccc}
    \toprule
         & \multicolumn{2}{c}{\textbf{zh}} & \multicolumn{2}{c}{\textbf{ko}} & \multicolumn{2}{c}{\textbf{th}} & \multicolumn{2}{c}{\textbf{ar}} & \multicolumn{2}{c}{\textbf{vi}} \\
\cmidrule{2-11}         & \textbf{general} & \textbf{specific} & \textbf{general} & \textbf{specific} & \textbf{general} & \textbf{specific} & \textbf{general} & \textbf{specific} & \textbf{general} & \textbf{specific} \\
    \midrule
    \textbf{GPT-4o-2024-11-20} & \textcolor[rgb]{ 0,  1,  0}{95.28} & \textcolor[rgb]{ 0,  0,  1}{27.78} & \textcolor[rgb]{ 1,  0,  0}{87.76} & \textcolor[rgb]{ 1,  0,  0}{45.16} & \textcolor[rgb]{ 0,  1,  0}{86.89} & \textcolor[rgb]{ 1,  0,  0}{35.00} & \textcolor[rgb]{ 0,  1,  0}{89.51} & \textcolor[rgb]{ 1,  0,  0}{42.42} & \textcolor[rgb]{ 1,  0,  0}{94.21} & \textcolor[rgb]{ 1,  0,  0}{57.14} \\
    \midrule
    \textbf{GPT-4o-mini} & 74.02 & 9.78 & 74.17 & \textcolor[rgb]{ 0,  0,  1}{29.27} & 67.79 & 16.84 & 72.84 & 19.82 & 80.17 & 35.16 \\
    \midrule
    \textbf{o1-mini} & 87.80 & 14.63 & \textcolor[rgb]{ 0,  0,  1}{83.80} & \textcolor[rgb]{ 0,  0,  1}{29.27} & 78.57 & 17.74 & 80.89 & 13.56 & 85.83 & 21.92 \\
    \midrule
    \textbf{Llama-3.1-Instruct-70B} & 87.20 & 11.96 & 78.15 & 23.17 & \textcolor[rgb]{ 0,  0,  1}{80.19} & 17.44 & 78.26 & 13.91 & 85.95 & 33.71 \\
    \midrule
    \textbf{Llama-3.1-Instruct-8B} & 46.03 & 5.56 & 42.38 & 14.10 & 52.66 & 11.70 & 49.37 & 4.26 & 69.42 & 20.69 \\
    \midrule
    \textbf{Qwen2.5-Instruct-72B} & 88.19 & 25.61 & 80.14 & 24.59 & 77.27 & 14.12 & 75.93 & 11.58 & 87.60 & 35.37 \\
    \midrule
    \textbf{Qwen2.5-Instruct-7B} & 57.94 & 19.64 & 40.85 & 11.90 & 45.81 & 10.00 & 35.22 & 7.53 & 60.50 & 23.19 \\
    \midrule
    \textbf{QwQ-32B} & 87.40 & 23.86 & 77.18 & 20.25 & 77.78 & 12.90 & 79.50 & 12.39 & 84.17 & 26.97 \\
    \midrule
    \textbf{QwQ-32B-Preview} & 80.99 & 26.92 & 78.68 & 26.83 & 79.43 & \textcolor[rgb]{ 0,  1,  0}{29.03} & 83.80 & 17.78 & 84.68 & 26.32 \\
    \midrule
    \textbf{QwQ-Plus} & 87.50 & 20.00 & 76.92 & 27.66 & 75.24 & 14.55 & 82.11 & 12.86 & 85.29 & 25.49 \\
    \midrule
    \textbf{Deepseek\_V3} & \textcolor[rgb]{ 0,  0,  1}{94.49} & \textcolor[rgb]{ 0,  1,  0}{35.96} & 82.12 & \textcolor[rgb]{ 0,  1,  0}{31.58} & 79.33 & \textcolor[rgb]{ 0,  0,  1}{20.00} & \textcolor[rgb]{ 0,  0,  1}{86.42} & \textcolor[rgb]{ 0,  0,  1}{23.42} & \textcolor[rgb]{ 0,  0,  1}{92.56} & \textcolor[rgb]{ 0,  0,  1}{46.51} \\
    \midrule
    \textbf{Deepseek\_R1} & \textcolor[rgb]{ 1,  0,  0}{96.85} & \textcolor[rgb]{ 1,  0,  0}{53.26} & \textcolor[rgb]{ 0,  1,  0}{86.75} & 25.61 & \textcolor[rgb]{ 1,  0,  0}{88.41} & \textcolor[rgb]{ 0,  0,  1}{20.00} & \textcolor[rgb]{ 1,  0,  0}{90.74} & \textcolor[rgb]{ 0,  1,  0}{25.22} & \textcolor[rgb]{ 0,  1,  0}{93.39} & \textcolor[rgb]{ 0,  1,  0}{52.17} \\
    \bottomrule
    \end{tabular}%
    \end{adjustbox}
    \caption{Model performance (CGA) on KoLasSimpleQA (part1/2).}
  \label{tab:res_CGA_1}%
\end{table}%

\begin{table}[htbp]
    \centering
    \begin{adjustbox}{width=\textwidth}
        \begin{tabular}{ccccccccccc}
    \toprule
         & \multicolumn{2}{c}{\textbf{cs}} & \multicolumn{2}{c}{\textbf{hu}} & \multicolumn{2}{c}{\textbf{ru}} & \multicolumn{2}{c}{\textbf{sr}} & \multicolumn{2}{c}{\textbf{avg.}} \\
\cmidrule{2-11}         & \textbf{general} & \textbf{specific} & \textbf{general} & \textbf{specific} & \textbf{general} & \textbf{specific} & \textbf{general} & \textbf{specific} & \textbf{general} & \textbf{specific} \\
    \midrule
    \textbf{GPT-4o-2024-11-20} & \textcolor[rgb]{ 0,  0,  1}{92.20} & \textcolor[rgb]{ 1,  0,  0}{42.86} & \textcolor[rgb]{ 1,  0,  0}{88.17} & \textcolor[rgb]{ 1,  0,  0}{32.00} & \textcolor[rgb]{ 0,  0,  1}{86.81} & \textcolor[rgb]{ 0,  0,  1}{27.91} & \textcolor[rgb]{ 1,  0,  0}{93.94} & \textcolor[rgb]{ 1,  0,  0}{46.27} & \textcolor[rgb]{ 0,  1,  0}{90.53} & \textcolor[rgb]{ 1,  0,  0}{41.89} \\
    \midrule
    \textbf{GPT-4o-mini} & 83.80 & 31.82 & 71.43 & 10.74 & 68.48 & 12.73 & 78.63 & 26.92 & 74.59 & 24.07 \\
    \midrule
    \textbf{o1-mini} & 86.76 & \textcolor[rgb]{ 0,  0,  1}{32.73} & 74.39 & 16.13 & 82.56 & \textcolor[rgb]{ 0,  1,  0}{31.03} & 86.15 & \textcolor[rgb]{ 0,  0,  1}{34.00} & 82.97 & 27.23 \\
    \midrule
    \textbf{Llama-3.1-Instruct-70B} & 89.44 & 26.97 & \textcolor[rgb]{ 0,  0,  1}{80.47} & 17.21 & 83.52 & 16.07 & \textcolor[rgb]{ 0,  0,  1}{91.73} & 32.43 & 83.88 & 24.01 \\
    \midrule
    \textbf{Llama-3.1-Instruct-8B} & 60.28 & 24.14 & 46.15 & 10.09 & 43.96 & 7.55 & 65.41 & 17.11 & 52.85 & 15.20 \\
    \midrule
    \textbf{Qwen2.5-Instruct-72B} & 81.69 & 22.78 & 52.69 & 7.37 & 72.53 & 22.45 & 73.68 & 17.14 & 76.64 & 22.55 \\
    \midrule
    \textbf{Qwen2.5-Instruct-7B} & 51.88 & 9.86 & 37.33 & 0.00 & 33.72 & 7.89 & 34.11 & 17.74 & 44.15 & 14.86 \\
    \midrule
    \textbf{QwQ-32B} & 83.10 & 26.51 & 63.86 & 12.07 & 78.26 & 19.64 & 81.95 & 20.51 & 79.24 & 21.85 \\
    \midrule
    \textbf{QwQ-32B-Preview} & 83.08 & 26.83 & 70.83 & \textcolor[rgb]{ 0,  1,  0}{19.05} & 85.19 & \textcolor[rgb]{ 1,  0,  0}{38.89} & 81.97 & \textcolor[rgb]{ 0,  1,  0}{37.84} & 80.96 & 30.29 \\
    \midrule
    \textbf{QwQ-Plus} & 85.71 & 15.56 & 66.33 & 9.09 & 77.59 & 17.50 & 82.54 & 22.22 & 79.91 & 20.71 \\
    \midrule
    \textbf{Deepseek\_V3} & \textcolor[rgb]{ 1,  0,  0}{92.96} & \textcolor[rgb]{ 0,  1,  0}{35.63} & 79.88 & 17.09 & \textcolor[rgb]{ 0,  1,  0}{87.91} & 16.98 & 88.72 & 29.49 & \textcolor[rgb]{ 0,  0,  1}{87.15} & \textcolor[rgb]{ 0,  0,  1}{31.09} \\
    \midrule
    \textbf{Deepseek\_R1} & \textcolor[rgb]{ 0,  1,  0}{92.86} & 31.46 & \textcolor[rgb]{ 0,  1,  0}{88.10} & \textcolor[rgb]{ 0,  0,  1}{17.60} & \textcolor[rgb]{ 1,  0,  0}{92.39} & 14.29 & \textcolor[rgb]{ 1,  0,  0}{93.94} & 24.36 & \textcolor[rgb]{ 1,  0,  0}{91.49} & \textcolor[rgb]{ 0,  1,  0}{31.14} \\
    \bottomrule
    \end{tabular}%
    \end{adjustbox}
    \caption{Model performance (CGA) on KoLasSimpleQA (part2/2).}
  \label{tab:res_CGA_2}%
\end{table}%

\begin{table}[h]
\centering
\small
\begin{tabular}{llllll}
\hline
\multicolumn{1}{c}{\multirow{2}{*}{Model}} & \multicolumn{2}{c}{language-specific domain} & \multicolumn{2}{c}{general domain} & \multicolumn{1}{c}{\multirow{2}{*}{mECE}} \\ \cline{2-5}
\multicolumn{1}{c}{}                       & tran\_en               & direct              & tran\_en          & direct         & \multicolumn{1}{c}{}                      \\ \hline
GPT-4o-2024-11-20                          & 0.42                   & 0.43                & 0.04              & 0.05           & 0.24                                      \\
GPT-4o-mini                                & 0.56                   & 0.55                & 0.07              & 0.14           & 0.33                                      \\
o1-mini                                    & 0.46                   & 0.48                & 0.07              & 0.1            & 0.28                                      \\
Llama-3.1-Instruct-70B                     & 0.42                   & 0.57                & 0.04              & 0.07           & 0.28                                      \\
Llama-3.1-Instruct-8B                      & 0.53                   & 0.63                & 0.11              & 0.3            & 0.39                                      \\
Qwen2.5-Instruct-72B                       & 0.61                   & 0.62                & 0.05              & 0.15           & 0.36                                      \\
Qwen2.5-Instruct-7B                        & 0.67                   & 0.71                & 0.15              & 0.38           & 0.48                                      \\
QwQ-32B                                    & 0.45                   & 0.43                & 0.04              & 0.1            & 0.26                                      \\
QwQ-32B-Preview                            & 0.32                   & 0.35                & 0.04              & 0.06           & 0.19                                      \\
QwQ-Plus                                   & 0.44                   & 0.43                & 0.05              & 0.09           & 0.25                                      \\
Deepseek\_V3                               & 0.51                   & 0.51                & 0.03              & 0.06           & 0.28                                      \\
Deepseek\_R1                               & 0.34                   & 0.33                & 0.04              & 0.06           & 0.19                                      \\
AvgECE                               & 0.48                   & 0.5                 & 0.06              & 0.13           & 0.29                                      \\ \hline
\end{tabular}
\caption{LLMs' Expected Calibration Error (ECE) in the general and language-specific domains corresponding to the \texttt{tran\_en} and \texttt{direct} settings. mECE represents the Mean of ECE across the two domains and two settings, AvgECE represents the average across the 12 LLMs.}
\label{tab:app_all_ece_result}
\end{table}

\section{Details of Analyzing the Reasoning Process of LRMs}
\label{sec:app_LRM_reasoning_analysis}

\begin{figure}[t]
  \centering
  \includegraphics[width=1.0\linewidth]{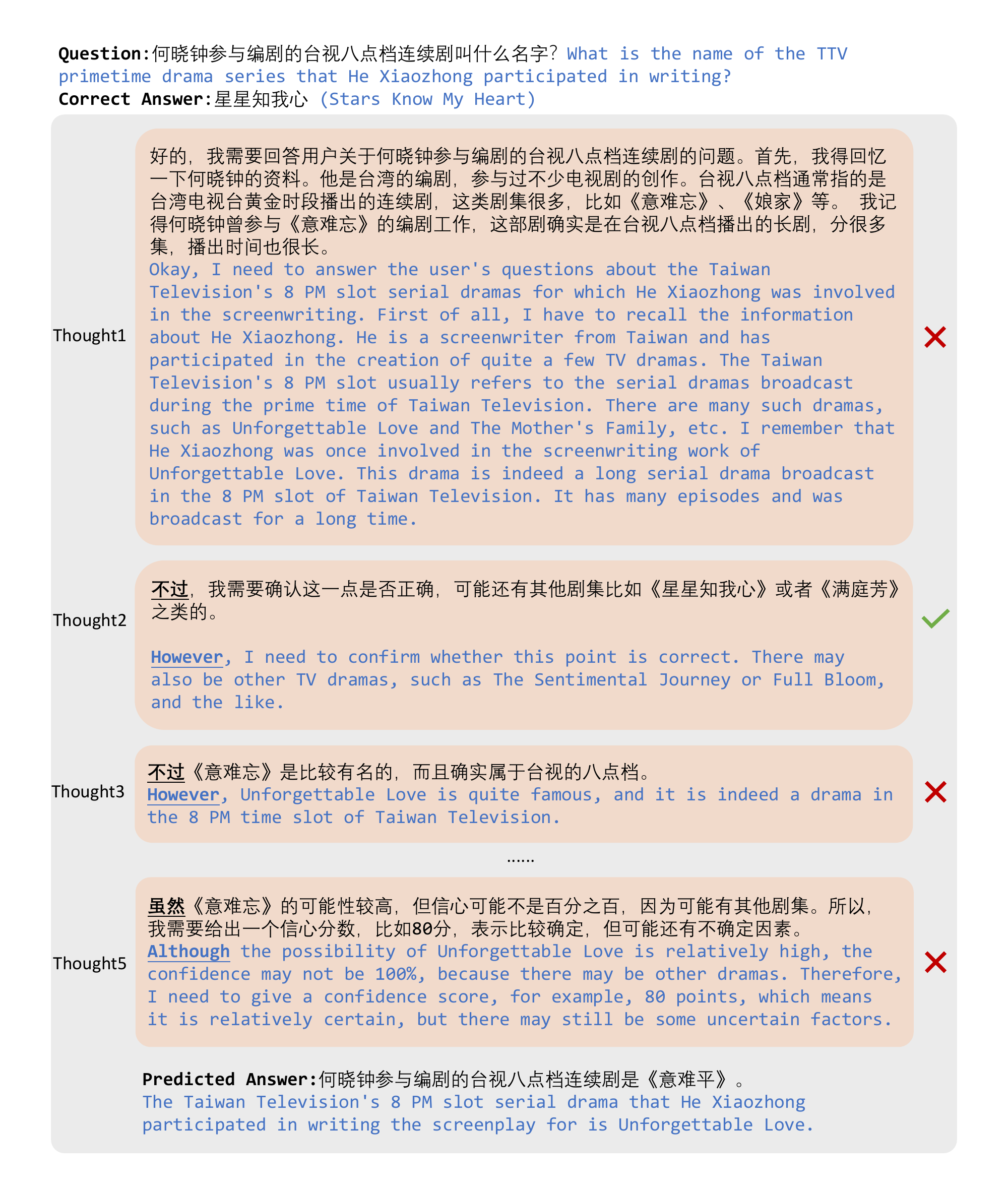}
  \caption{
  Example of the thoughts in Deepseek-R1’s reasoning process on KoLasSimpleQA. \textcolor{green}{\ding{51}} and \textcolor{red}{\ding{55}} indicate the
 correctness of the thoughts. The original text is in black, while the translation into English is in \textcolor{blue}{blue}.
  }
  \label{fig:app_ds_reasoning_example}
\end{figure}

\begin{figure}[t]
  \centering
  \includegraphics[width=1.0\linewidth]{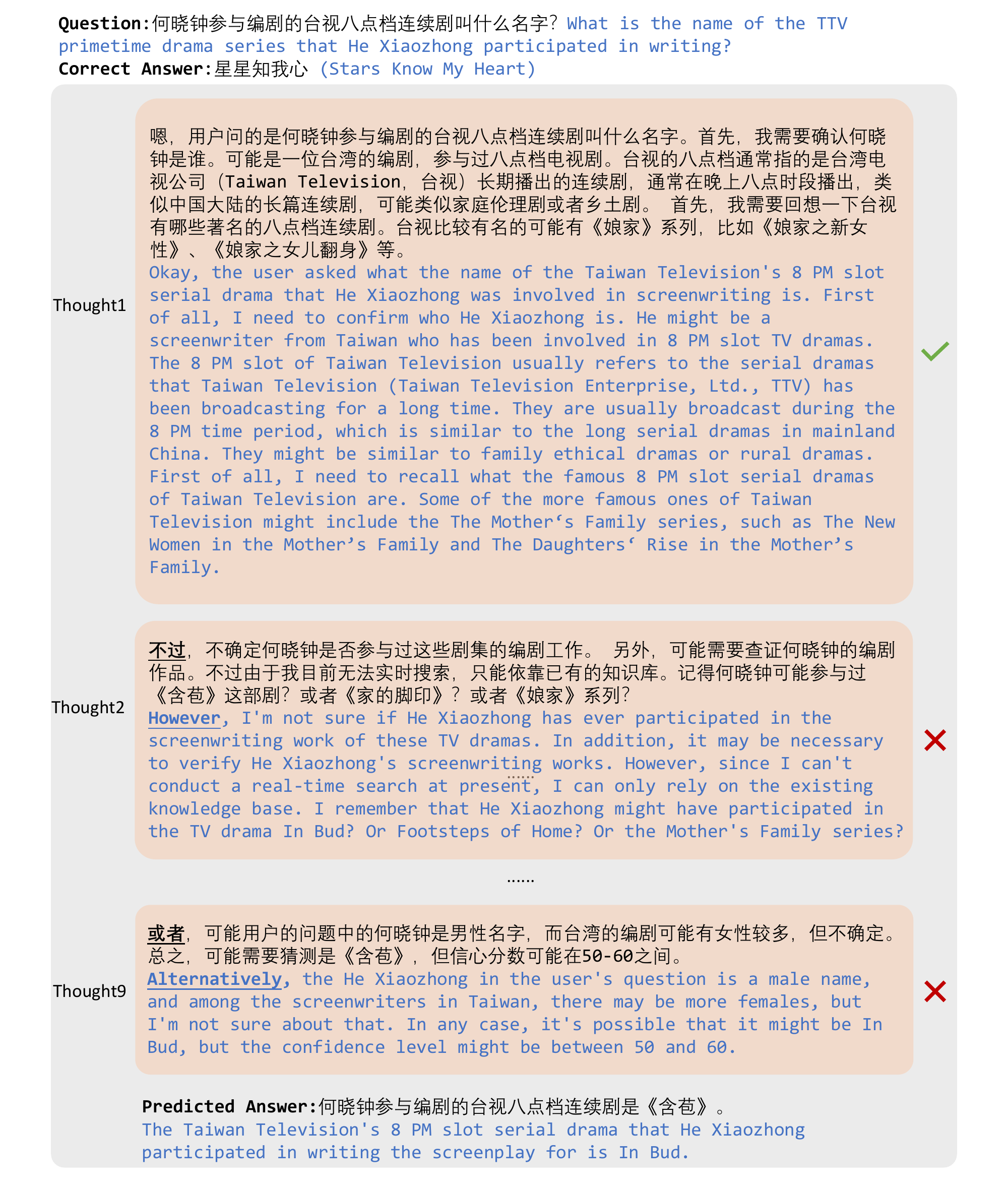}
  \caption{
    Example of the thoughts in QwQ-32B’s reasoning process on KoLasSimpleQA. \textcolor{green}{\ding{51}} and \textcolor{red}{\ding{55}} indicate the
 correctness of the thoughts. The original text is in black, while the translation into English is in \textcolor{blue}{blue}.
  }
  \label{fig:app_qwq_reasoning_example}
\end{figure}

We employed GPT-4o to segment the reasoning process into distinct thoughts (see the prompts in Table~\ref{tab:app_LRM_reasoning_analysis_prompt_shift_expression_extracting} and~\ref{tab:app_LRM_reasoning_analysis_prompt_shift_expression_confirming}). Furthermore, we assessed the correctness of each thought (see the prompt in table \ref{tab:app_LRM_reasoning_analysis_prompt_thought_access}), with additional examples provided in Figure~\ref{fig:app_ds_reasoning_example} and~\ref{fig:app_qwq_reasoning_example}.

\begin{table}[t]
\centering
\small
\begin{tcolorbox}[title={Shift Expression Extracting Prompt}, colback=white, coltitle=black, colbacktitle=white!0]

Given a thinking process for answering a question, follow these steps to extract contrastive expressions from the answer text:\\
1.	Identify the Primary Language:\\
•	First, determine the primary language of the answer text.\\
2.	Extract Contrastive Words, Phrases, or Expressions:\\
	•	Identify all the phrases that express a shift in opinion, explanation, or answer, phrases that signal a contrast or change in direction.\\ 
	•	For English: “However,” “but,” “On the other hand,” “Although,” “Nevertheless,” “Yet,” “Despite,” “In contrast,” “Instead,” “Even though.”\\
	•	Please pay attention that phrases indicating a successive relationship, such as "so", "for example" and the like, must never appear in your answers. Your goal is to find phrases indicating a contrast of viewpoints.\\

Requirements:\\
	1.	Identify and list all the contrastive words or phrases that indicate a shift in meaning, thought, or direction.\\
	2.	These expressions should be at the beginning of a sentence to signal a shift.\\
	3.	Keep the original text’s meaning and context intact.\\
	4.	Ensure to maintain the original capitalization of the words (e.g., “However” vs. “however”).\\
	5.	Provide a clear list of all the identified contrast words or phrases.\\

\textbf{[Input text]:}
\begin{verbatim}
{
    "question": <question>,
    "answer": <answer>,
}
\end{verbatim}

Please respond strictly in JSON format. Do not include any additional text outside the JSON structure. The output should also include the detected language type.\\

\textbf{[Output]:}
\begin{verbatim}
{
    "language": "<detected_language>",
    "shift_expression":[list]
}
\end{verbatim}

\end{tcolorbox}
\caption{Prompt template for splitting the LRM’s reasoning process into thoughts on KoLasSimpleQA (step 1/2)}
\label{tab:app_LRM_reasoning_analysis_prompt_shift_expression_extracting}
\end{table}
\begin{table}[t]
\centering
\small
\begin{tcolorbox}[title={Shift Expression Confirming Prompt}, colback=white, coltitle=black, colbacktitle=white!0]

Given the thinking process, identify all the phrases that express a shift in opinion, explanation, or answer, i.e., phrases that signal a contrast or change in direction (commonly known as “contradiction,” “contrast,” or “transition” phrases). For each identified phrase, wrap it in the format 
\verb|<shift_phrase_X>word<shift_phrase_X>|, where X is the sequential number for each occurrence of the phrase.\\
You should provide a list of the sequence numbers corresponding to the phrases that convey a shift in meaning.\\
If there is no \verb|"<shift_phrase_X>"| tag in text, return an empty list.\\

\textbf{[Input text]:}\verb|<input_text>|\\

Please respond strictly in JSON format. Do not include any additional text outside the JSON structure.\\ 

\textbf{[Output]:}
\begin{verbatim}
{
    "shift_phrase_ids": [list of sequential numbers of shifting phrases]
}
\end{verbatim}

\end{tcolorbox}
\caption{Prompt template for splitting the LRM’s reasoning process into thoughts on KoLasSimpleQA (step 2/2)}
\label{tab:app_LRM_reasoning_analysis_prompt_shift_expression_confirming}
\end{table}
\begin{table}[t]
\centering
\small
\begin{tcolorbox}[title={Thought Access Prompt}, colback=white, coltitle=black, colbacktitle=white!0]

Problem P = \verb|<problem>|\\ 
Solution Draft S = \verb|<split_solutions>|\\ 
Correct Answer A = \verb|<expected_answer>|\\

1. Please analyze the relevance between the solution S and the problem P, and conduct some verifications to check the correctness of the solution itself. Please think step by step to give an explanation **EXPLANATION**.\\
2. If you think the solution draft S can lead to the correct answer A of the problem P, please stick to the line of thinking without deviation and carry it through to completion. If you think it cannot yield the correct answer or you’re not sure, indicate a confidence score of 0 out of 2.\\
3. Please tell me honestly how confident you are that you can solve the problem P correctly based on the the solution draft S. Out of 2, please generate your confidence score **CONFIDENT SCORE**.\\

Please output **EXPLANATION** and **CONFIDENT SCORE** according to the following JSON format:
\begin{verbatim}
{
    "explanation": "EXPLANATION",
    "confident_score": "CONFIDENT SCORE"
}
\end{verbatim}

\end{tcolorbox}
\caption{Prompt template for evaluating the correctness of the thoughts of the LRM’s reasoning process on KoLasSimpleQA.}
\label{tab:app_LRM_reasoning_analysis_prompt_thought_access}
\end{table}

\end{CJK}
\end{document}